\documentclass[twocolumn, natbib]{svjour3}

\RequirePackage{fix-cm}
\usepackage{graphicx}%
\usepackage{multirow}%
\usepackage{amsmath,amssymb,amsfonts}%

\usepackage{mathrsfs}%
\usepackage[title]{appendix}%
\usepackage[pagebackref=true,breaklinks=true,citecolor=blue,linkcolor=blue,colorlinks]{hyperref}
\usepackage{xcolor}%
\usepackage{textcomp}%
\usepackage{manyfoot}%
\usepackage{booktabs}%
\usepackage{algorithm}%
\usepackage{algorithmicx}%
\usepackage{algpseudocode}%
\usepackage{listings}%
\usepackage{enumerate}

\smartqed

\usepackage{bm}  
\usepackage{xspace} 
\usepackage{booktabs}
\usepackage{adjustbox}
\usepackage{makecell}
\usepackage{mwe}

\usepackage{anyfontsize}

\newcolumntype{R}[2]{%
    >{\adjustbox{angle=#1,lap=1.3\width-(#2)}\bgroup}%
    l%
    <{\egroup}%
}

\begin{document}\sloppy


\newcommand{\systemname}{{OA-DET3D}\xspace}
\newcommand{\parsection}[1]{\vspace{4pt}\noindent\textbf{#1:}}
\newcommand{\psection}[1]{\vspace{4pt}\noindent\textbf{#1\\}}

\title{\systemname: Embedding Object Awareness as a General Plug-in for Multi-Camera 3D Object Detection}

\author{Xiaomeng Chu$^1$ \and Jiajun Deng$^2$ \and Jianmin Ji$^1$ \and Yu Zhang$^1$ \and Houqiang Li $^1$ \and Yanyong Zhang$^1$}
\authorrunning{Xiaomeng Chu~\etal} 

\institute{
	Xiaomeng Chu \at
	\email{cxmeng@mail.ustc.edu.cn}           
	\and
	Jiajun Deng, corresponding author \at
	\email{djiajun1206@gmail.com}
    \and
    Jianmin Ji \at
	\email{jianmin@ustc.edu.cn}
	\and
	Yu Zhang \at
	\email{yuzhang@ustc.edu.cn}
	\and
	Houqiang Li \at
	\email{lihq@ustc.edu.cn}
	\and
	Yanyong Zhang, corresponding author \at
	\email{yanyongz@ustc.edu.cn}
	\\
 \\
$^1$University of Science and Technology of China, Hefei, Anhui, China\\
\\
$^2$The University of Sydney, Sydney, Australia\\
%
}

\date{Received: date / Accepted: date}






\maketitle
\begin{abstract}
The recent advance in multi-camera 3D object detection is featured by bird's-eye view (BEV) representation or object queries.
However, the ill-posed transformation from image-plane view to 3D space inevitably causes feature clutter and distortion, making the objects blur into the background. 
To this end, we explore how to incorporate supplementary cues for differentiating objects in the transformed feature representation. 
Formally, we introduce \systemname, a general plug-in module that improves 3D object detection by bringing object awareness into a variety of existing 3D object detection pipelines.
Specifically, \systemname boosts the representation of objects by leveraging object-centric depth information and foreground pseudo points.
First, we use object-level supervision from the properties of each 3D bounding box to guide the network in learning the depth distribution. Next, we select foreground pixels using a 2D object detector and project them into 3D space for pseudo-voxel feature encoding. Finally, the object-aware depth features and pseudo-voxel features are incorporated into the BEV representation or query feature from the baseline model with a deformable attention mechanism.
We conduct extensive experiments on the nuScenes dataset and Argoverse 2 dataset to validate the merits of \systemname. Our method achieves consistent improvements over the  BEV-based baselines in terms of both average precision and comprehensive detection score.

\end{abstract}



\section{Introduction}
\label{sec:intro}

\begin{figure}[t]
    \centering
    \includegraphics[width=0.96\columnwidth]{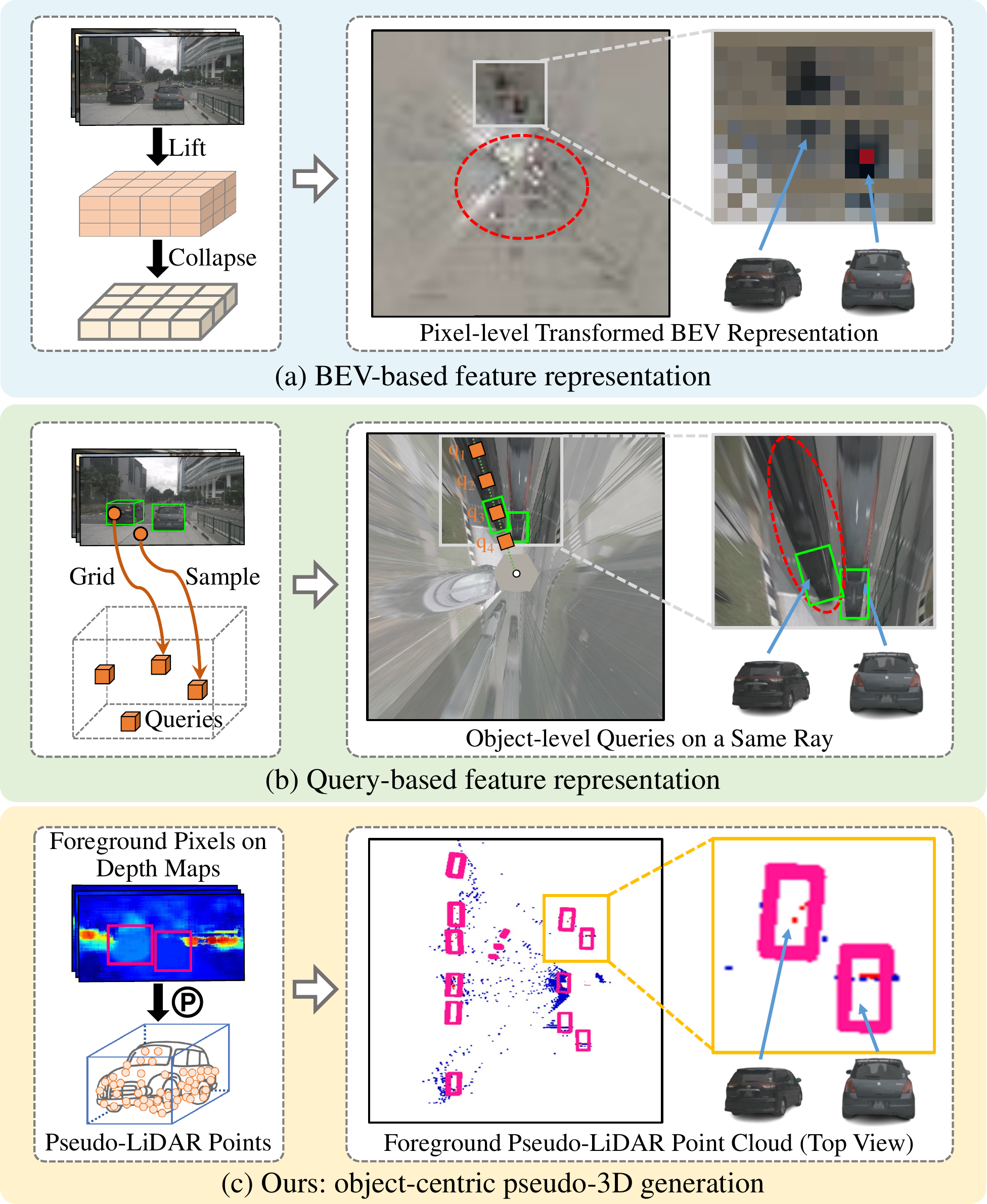}
    \caption{Comparison and visualization of different view transformation methods. (a) The BEV-based feature generation method. (b) The query-based feature generation method. (c) The object-centric pseudo-3D generation method.} 
    \label{fig:comp}
    \vspace{-0.2cm}
\end{figure}

Multi-view cameras have emerged as a popular choice of sensors for 3D object detection in autonomous driving, as evidenced by recent studies~\cite{3dod_ijcv_survey1, 3dod_ijcv_survey2, 3dod_tpami_survey, auto_driving_survey1}. This is mainly due to their comprehensive perception capabilities across the entire field of view, available at a relatively low cost. 
The exploration of multi-camera 3D object detection is primarily along two distinct directions: BEV-based methods and query-based methods.

BEV-based methods transform multiple camera views into a unified bird’s eye view, simplifying the understanding of spatial relationships and distances between objects. 
The existing view transformation methods either lift image pixels to a 3D frustum following the geometric principles and then splat them onto the BEV plane~\cite{LSS, BEVDet, BEVDepth}, or form the dense grid-shaped BEV queries by sampling image features via attention mechanisms~\cite{BEVFormer, Attention, Deformable-DETR}.
However, these methods treat all pixels equally, regardless of whether a pixel belongs to an object or not. No particular attention is given to pixels that belong to objects of interest.
As shown in Figure~\ref{fig:comp}(a), due to the pixel-level image features being transformed without distinguishing between the foreground and background, the resulting BEV features are noisy, obscuring the objects' locations from clear background differentiation.
Inaccurate BEV features, particularly those describing objects, can significantly compromise the detection ability of the BEV head.

On the contrary, query-based 3D object detectors directly extract query features from multi-camera inputs without explicit view transformation.
Typically, these detectors use a set of learnable parameters or uniform grid-like distributions in 3D space as the initial object queries~\cite{DETR3D, sparsebev}, projecting each query onto the image plane to sample features from specific locations.
However, query-based feature sampling also has its issues. 
Without prior knowledge of depth, queries along each camera ray, such as q1 to q4 highlighted in Figure \ref{fig:comp}(b), sample the same image location, resulting in considerable ambiguity regarding the object's position along the depth axis.
To better visualize, we evenly distribute dense queries on the BEV map, it can be noticed that the features of the object form a noticeable tailing phenomenon on the BEV, as illustrated by the red circle in Figure \ref{fig:comp}(b).

In this paper, our objective is to provide additional cues for distinguishing objects from background information by leveraging explicit depth features and object-centric 3D features. First, we estimate depth under object-level supervision. Second, we localize objects in images and project the pixels belonging to objects into a foreground pseudo-point cloud, (\emph{e.g.} object-centric pseudo-3D representation), to extract voxel features. These two features provide valuable 3D spatial information for each object. By integrating them with BEV features or query features obtained from the baseline model, we enhance the spatial details of objects, thereby rendering the detection system more \emph{object aware}. Formally, we introduce \systemname, which incorporates a branch designed to enhance BEV-based 3D object detection by harnessing these two object-aware features as explained above.

The main challenge in generating the two object-aware features is to avoid using any additional depth labels. To address this, we use either the center point or multiple points along the diagonal of an object's 3D bounding box to supervise foreground pixels in multi-camera images and obtain the estimated depth map. We then employ a 2D detector to predict 2D bounding boxes and select the corresponding foreground regions on the depth maps. Next, we project these selected foreground depth map pixels into 3D space to form sparse pseudo-LiDAR point clouds~\cite{Pseudo-Lidar}, 
which are indeed sparse but give a more accurate representation of the object location than the BEV representation, as demonstrated in Figure~\ref{fig:comp}(c).
After voxelization and multiple layers of 3D sparse convolutions~\cite{sparse-3D-conv}, we encode the foreground pseudo-LiDAR point clouds as pseudo-voxel features. Finally, \systemname fuses these features with the depth features using deformable attention.

To demonstrate the effectiveness of our proposed \systemname, we evaluate its performance on the challenging nuScenes benchmark~\cite{nuScenes} and Argoverse 2 dataset~\cite{av2}. Plugged in with two BEV-based baseline networks BEVDet~\cite{BEVDet} and BEVFormer~\cite{BEVFormer} and two query-based baseline networks SparseBEV~\cite{sparsebev} and StreamPETR~\cite{streampetr}, \systemname consistently improves detection performance. 
OA-BEVFormer outperforms BEVFormer on the nuScenes validation set, achieving a 43.1\% mAP and 52.8\% NDS, which represents a respective increase of 1.5\% and 1.1\%. Impressively, on the test set, OA-BEVFormer delivers a robust performance of 49.4\% mAP and 57.5\% NDS. Furthermore, OA-StreamPETR also shows remarkable results, scoring a 55.5\% mAP on the nuScenes test set.

In summary, our main contributions are as follows:
\begin{itemize}
    \item 
    We propose \systemname, a plug-in module that brings object awareness to the transformed feature representation from the image-plane view for multi-camera 3D object detection, thereby compensating for feature clutter and deformation in view transformation module.
    \item We generate object-aware features by reusing 3D object detection labels, without introducing extra depth annotation. 
    Such lightweight operations can lead to considerable improvements. 
    \item We conduct experiments with two representative BEV-based baselines, \emph{i.e.}, BEVFormer and BEVDet, and two representative query-based baselines, \emph{i.e.}, SparseBEV and StreamPETR, with multiple backbones on the nuScenes dataset and Argoverse 2 dataset. With \systemname plugged in, the performance improvements in both methods verify the utility of our approach. 
\end{itemize}

\section{Related work}
\label{sec:related_work}

In this section, we give a brief review of visual 3D object detection, including monocular 3D object detection and multi-camera 3D object detection.
The latter can be broadly categorized into two approaches: BEV-based~\cite{BEVDepth, PolarFormer, BEVerse, BEVDet4D, hop} and query-based~\cite{DETR3D, PETR, sparse4dv3, mv2d, far3d}.

\begin{figure*}[t]
    \centering
    \includegraphics[width=0.95\textwidth]{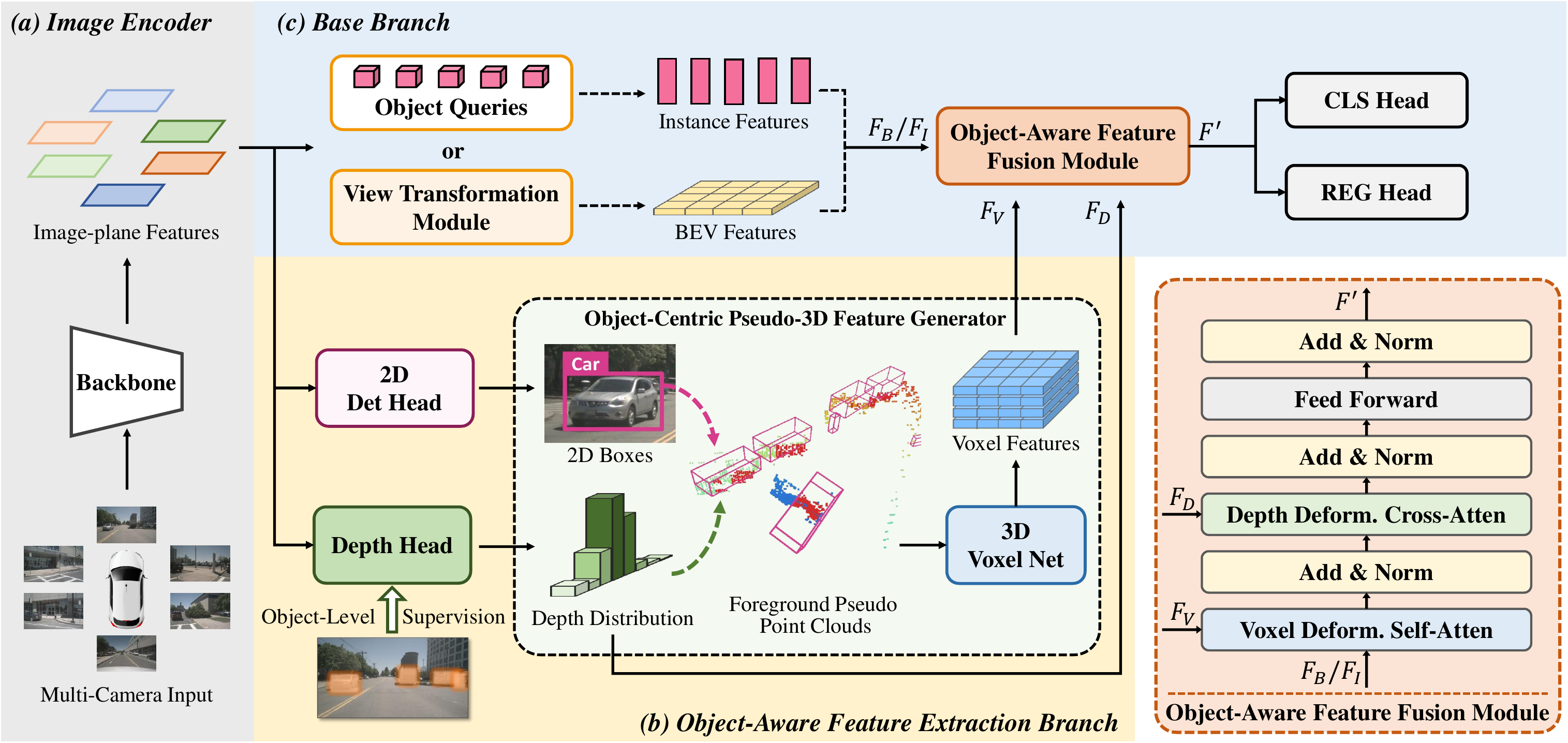}
    \caption{
    The overall architecture of \systemname comprises (a) an image encoder, (b) an object-aware feature extraction branch, and (c) a base branch. Branch (b) consists of a 2D detection head, a depth head with object-level supervision, and an object-centric pseudo-3D feature generator. Meanwhile, branch (c) includes a view transformation module for BEV-based pipelines or pre-defined queries for query-based pipelines, an object-aware feature fusion module, a classification head and a regression head. In this framework, the object-aware features $F_V$ and $F_D$ are extracted in branch (b), and subsequently fused with the base features $F_B$ or $F_I$ via the object-aware feature fusion module to obtain the updated features $F'$.
    }  
    \label{fig:pipeline}
    \vspace{-0.1cm}
\end{figure*}

\parsection{Monocular 3D Object Detection}
In the context of monocular image inputs, two types of 3D object detection methods exist. 
The first category supervises networks solely based on attributes specific to a 3D object~\cite{Deep3DBox, MonoDETR, M3D-RPN, RTM3D, MonoFlex}, while the second type leverages additional data such as depth or semantic labels\cite{AM3D, patchnet, CaDDN, Neighbor-Vote}. We provide examples of each below. 
FCOS3D~\cite{FCOS3D} is a single-stage 3D object detector that infers the 7-degree-of-freedom properties of 3D objects in images. 
SMOKE~\cite{SMOKE}, on the other hand, employs a single keypoint estimate and regressed 3D variables for eight bounding box corners to construct 3D bounding boxes. 
D4LCN~\cite{D4LCN} introduced a unique local convolutional network where filters are generated dynamically from the depth map.
Finally, Y. Wang et al.~\cite{Pseudo-Lidar} introduced ``Pseudo-LiDAR", a representation obtained by combining image pixel coordinates with corresponding depth value.

\parsection{BEV-based Multi-Camera 3D Object Detection}
BEVDet~\cite{BEVDet} constructs BEV features using the lifting-collapsing method and performs feature extraction in both image and BEV spaces. BEVFormer~\cite{BEVFormer} extracts spatial features across camera views into BEV through deformable cross-attention, while also aggregating temporal information to enhance detection performance. 
BEVStereo~\cite{BEVStereo} overcomes depth perception ambiguity by dynamically selecting matching candidates to reduce computation cost and using an iterative algorithm to adapt to moving objects.
FB-BEV~\cite{FB-BEV} addresses limitations in existing paradigms by combining the strengths of forward projection and backward projection, mutually enhancing and obtaining higher-quality BEV representations.
SOLOFusion~\cite{SOLOFusion} proposes a long-term temporal fusion approach for camera-only 3D detection that generates a cost volume from extensive image observations to optimize multi-view matching.
VideoBEV~\cite{videobev} proposes a long-term recurrent fusion strategy for BEV 3D perception that efficiently integrates history information, complemented by a temporal embedding module for robustness against missed frames.

\parsection{Query-based Multi-Camera 3D Object Detection}
In recent years, DETR3D~\cite{DETR3D} and PETR~\cite{PETR} have leveraged deformable transformers~\cite{Deformable-DETR} for this task, with the former utilizing 3D object queries to index 2D feature points, while the latter embeds 3D position information into 2D features. 
StreamPETR~\cite{streampetr} introduces an object-centric temporal mechanism and motion-aware layer normalization for multi-view 3D object detection, achieving significant performance improvements with minimal computation cost. 
SparseBEV~\cite{sparsebev} incorporates scale-adaptive self-attention for feature aggregation in BEV space, adaptive spatio-temporal sampling guided by queries, and adaptive mixing for dynamic feature decoding.
Sparse4D~\cite{sparse4d} achieves iterative refinement of anchor boxes through sparse sampling and fusion of spatial-temporal features. Additionally, it incorporates an instance-level depth reweight module to address the issues in 3D-to-2D projection.
MV2D~\cite{mv2d} harnesses 2D detection to generate 3D object queries informed by image semantics, enhancing detection and localization. It employs a sparse cross-attention module for precise feature focus and noise reduction.
RayDN~\cite{RayDN} boosts 3D object detection accuracy by creating depth-aware features through strategic sampling along camera rays, designed as a versatile module for DETR-style detectors with minimal impact on computational costs.

\section{Methodology}
\label{sec:methodology}

\subsection{Overview}

\noindent\textbf{3D Detection with \systemname:} Figure~\ref{fig:pipeline} depicts our \systemname pipeline. 
It can serve as a plugin to integrate object-aware information, including depth estimates with object-level supervision and foreground pseudo-3D features, into any BEV-based or query-based multi-camera 3D detection pipeline.
With \systemname plugged in, the resulting object-aware 3D detection usually consists of an image encoder followed by two branches: (1) the object-aware feature extraction branch and (2) the base branch. 
The former includes three modules: (i) 2D detection head, (ii) depth head, and (iii) object-centric pseudo-3D feature generator.
The latter can be any BEV-based or query-based baseline model. For BEV-based methods, the branch obtains BEV features through a view transformation module. For query-based methods, the branch projects object queries onto the image plane to obtain instance features. This is followed by an object-aware feature fusion module and a set of classification and 3D bounding box regression heads.

The workflow can be summarized as follows: 
first, we input multi-camera images into the backbone networks to extract image features. 
These features are forward-projected through the view transformation module to obtain BEV features, or sampled by instance-level queries to get instance features, and then fed into the 2D detection head to predict 2D bounding boxes, as well as the depth head to estimate depth maps with object-level supervision.
Second, the depth-map pixels selected by the 2D bounding boxes are projected into the 3D space to form foreground pseudo-point clouds, from which object-centric voxel features can be encoded by a 3D voxel network. The depth features and pseudo-voxel features are fused with the BEV or instance features using deformable attention modules. Finally, the enhanced features are fed into the classification head and regression head for 3D object detection.

\noindent\textbf{Design Overview of \systemname:} The key modules of the proposed
\systemname
are as follows: 
\begin{enumerate}[(1)]
\item Depth Estimation with Object-Level Supervision.
This module is to explicitly guide the network to learn the depth, as well as to avoid the need for extra annotations.
We leverage the properties of a 3D bounding box, such as its center point or diagonal, to approximate the depth of corresponding pixels within the 2D bounding box.
\item Object-Centric Pseudo-3D Feature Generation.
This module is to construct a foreground pseudo-3D representation that encodes features describing objects of interest in the 3D space. Pixels belonging to an object’s 2D bounding box on depth maps are selected and projected to 3D space to generate pseudo-point clouds, which undergo voxelization followed by encoding pseudo-voxel features through layers of 3D sparse convolutions. This process is trained end-to-end.

\item Object-Aware Feature Fusion.
This module is to fuse the two object-aware features into the BEV features or instance features from the based branch through deformable attention.

\end{enumerate}
In the rest of this section, we discuss these modules in detail.

\subsection{Depth Estimation with Object-Level Supervision}
\begin{figure}[t]
    \centering
    \includegraphics[width=0.95\columnwidth]{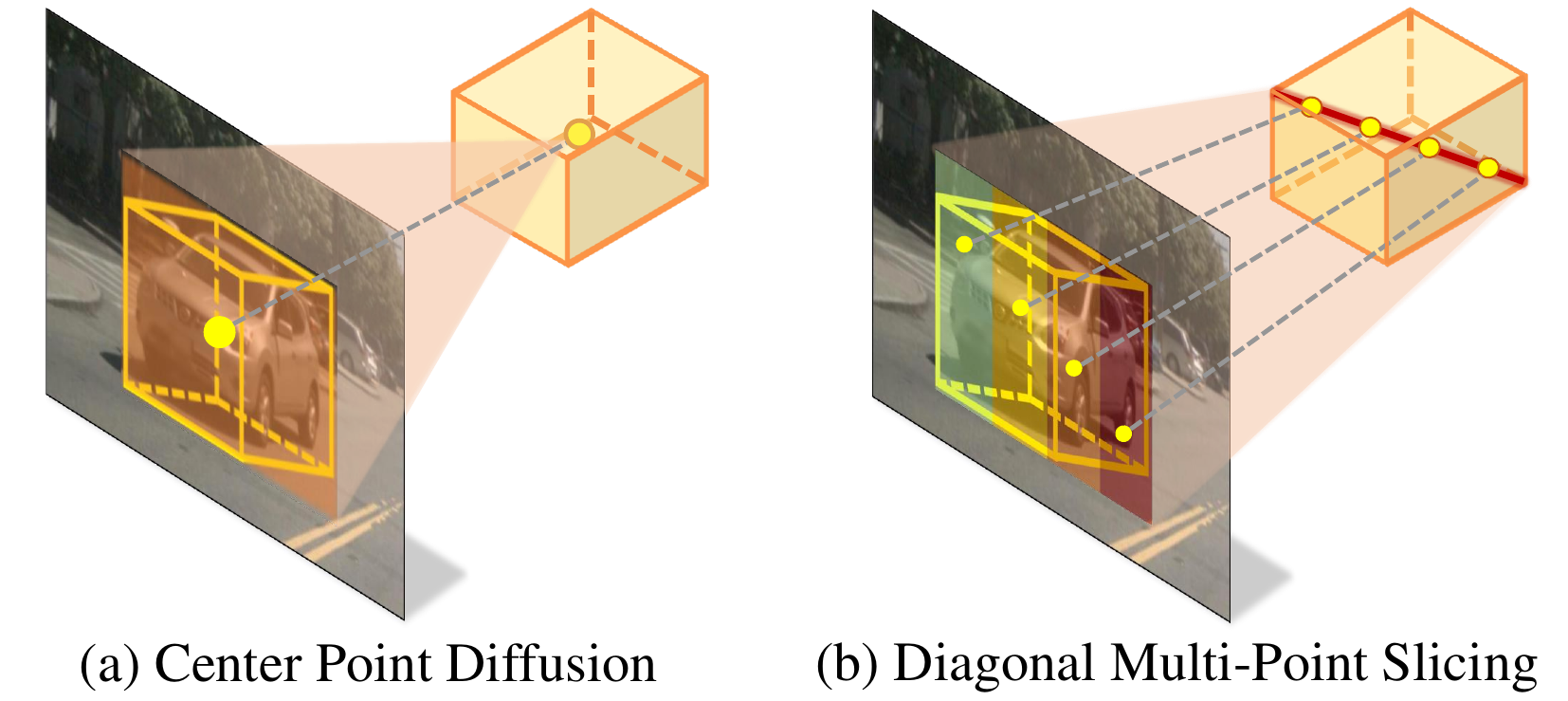}
    \caption{
    Comparison of two object-level depth supervision methods: center point diffusion and diagonal multi-point slicing. 
    The former assigns a uniform depth value to all pixels within the object's 2D bounding box, while the latter evenly selects $N$ points along the diagonal of a 3D bounding box to determine the depth value for each slice of its corresponding 2D bounding box.
    } 
    \label{fig:object-level_supervision}
    \vspace{-0.1cm}
\end{figure}

When depth labels from LiDAR point clouds are not available, learning depth information during view transformation is typically achieved implicitly~\cite{BEVDet}. 
In \systemname, we utilize two distinct object-level depth supervision methods: (a) center point diffusion and (b) diagonal multi-point slicing, as depicted in Figure~\ref{fig:object-level_supervision}. 
The former method assigns the depth value of a 3D object's center point to all pixels within the object's 2D bounding box~\cite{MonoDETR}. 
The latter method involves evenly selecting $N$ points along the diagonal of a 3D bounding box, projecting them onto the image plane, and subsequently vertically slicing the object's 2D bounding box based on these points. Here, the depth value of each point is used as the depth value for its corresponding slice. We will compare and analyze the effectiveness of these two methods in Section~\ref{sec:experiments}.
While such supervision yields only an approximation, it provides explicit supervision that proves crucial in the subsequent generation of pseudo-3D representations. To enhance prediction accuracy, we discretize the depth range into unequal intervals and treat each interval as a distinct class, thereby casting depth estimation as a classification task in an ordinal regression way~\cite{DORN}. Specifically, we utilize a linearly-increasing discretization (LID) strategy with a bin size of $\sigma$ to discretize the depth range $[d_{min}, d_{max}]$ as follows:
\begin{flalign}
    &&
    \begin{aligned}
    \delta &= \frac{2(d_{max}-d_{min})}{K(K+1)}, \\
    \hat{l} &= \lfloor-0.5+0.5\sqrt{1+\frac{8(\hat{d}-d_{min})}{\delta}}\rfloor,
    \end{aligned}
    &&
\end{flalign}
where $K$ denotes the number of depth bins, and $\hat{l}$ is the bin index.

As such, the depth estimation task is formulated as a classification problem with $K$ classes.  
Specifically, the depth head produces a feature map $y$ of size $H\times W\times K\times 2$, and subsequently, the depth probability distribution $P$ at coordinate $(w, h)$ is obtained by applying a softmax to $y_{(w, h)}$, in accordance with the following expression:
\begin{flalign}
    &&
    P^{l}_{(w,h)} = softmax(y^{l}_{(w,h)}),
    &&
\end{flalign}
where$P^{l}_{(w,h,0)}$ denotes the probability of the depth value lying within the interval $[l, K]$. The depth category $d_c$ is determined by counting intervals where probabilities exceed 0.5, and the final depth $d_{(w,h)}$ is assigned the median value of the $d_c$-th interval in meters.

\subsection{Object-Centric Pseudo-3D Feature Generation}

\begin{figure}[t]
    \centering
    \includegraphics[width=0.95\columnwidth]{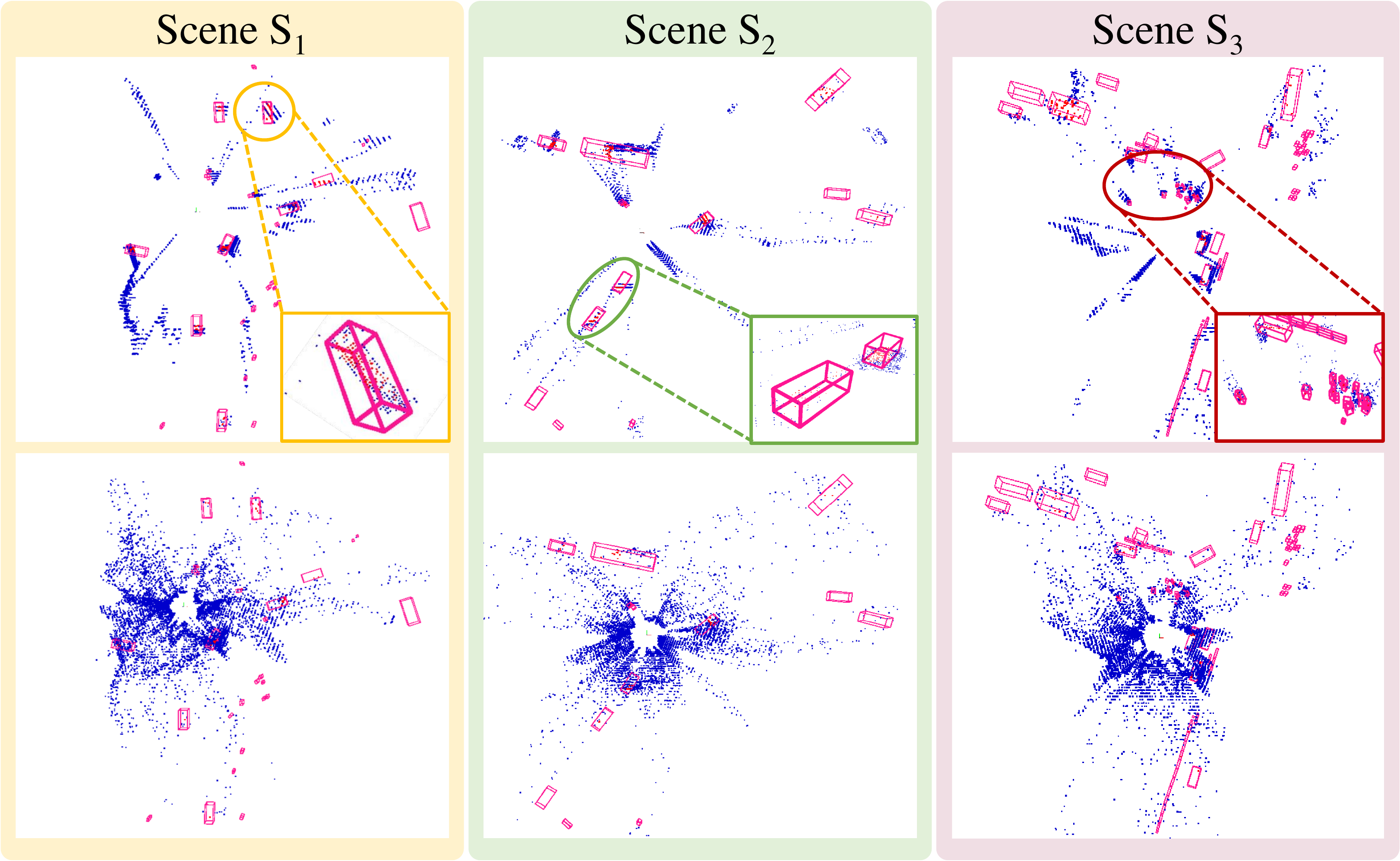}
    \caption{Comparison between foreground pseudo-LiDAR points (top) and all points, including background ones (bottom). Object-level depth supervision allows foreground pseudo-LiDAR point clouds to precisely depict object locations, while background points contain significant noise.    } 
    \label{fig:fg}
    \vspace{-0.1cm}
\end{figure}

This section discusses how to generate pseudo-3D features that provide critical object-aware 3D spatial features. After obtaining depth maps, we integrate the 2D object detector FCOS~\cite{FCOS} into our pipeline to predict 2D bounding boxes on multi-camera images. Specifically, after the image encoder, we incorporate an FPN~\cite{FPN} and a 2D object detection head into our implementation. Only pixels within these boxes are considered foreground pixels, and we restrict subsequent steps to focus solely on objects of interest. We combine the selected pixel coordinates $(u, v)$ with the corresponding depth to generate a pseudo-LiDAR point $(x, y, z)$. The transformation can be formulated as follows:
\begin{flalign}
    &&
    x = \frac{(u - c_x)z}{f_x}, \quad\!
    y = \frac{(v - c_y)z}{f_y}, \quad\!
    z = D(u, v),
    &&
\end{flalign}
where $(c_x, c_y)$ refers to the camera center in image pixel location, and $z$ denotes the estimated depth of the pixel produced by the depth head $D$.
Figure~\ref{fig:fg} illustrates that depth supervision derived from ground-truth boxes allows foreground pseudo-LiDAR point clouds to accurately determine object locations, while background points contain considerable noise. By removing background points, object-centric pseudo-point clouds provide superior descriptions of objects' spatial properties, such as location and 3D structure. We jointly train the 2D object detector and 3D object detection in an end-to-end manner.

To balance accuracy and efficiency, we employ a voxel-based 3D feature encoder instead of the point-based approach~\cite{pointnet,pointnet++,Voxel-R-CNN,PV-RCNN++}. Pseudo-point cloud voxelization results in sparse 4D tensors with dimensions $(H_\text{B}, W_\text{B}, Z, C)$ representing the 3D space. Voxel features $F_{V}$ are extracted through multi-layer 3D sparse convolutions. Flattening voxel features along the z-axis converts them into BEV features with the size of $(H_\text{B}, W_\text{B}, Z\cdot C)$.

\subsection{Object-Aware Feature Fusion}

Finally, it is necessary to aggregate the two types of object-aware features: the pseudo-3D feature $F_V$ (flattened into BEV) and the depth feature $F_D$, with the base features $F$ from the base branch. 
The transformer architecture exhibits significant superiority in the adaptive integration of distinct feature tensors, characterized by variations in both shape and semantics. This capability is attributed to its inherent flexibility and adaptability, which allows for the effective fusion of diverse data representations within a unified framework.
Moreover, deformable attention~\cite{Deformable-DETR} provides an effective means to establish soft associations between misaligned features. Given the positional and semantic misalignment between the object-aware features and features from the base branch, we utilize the deformable transformer to fuse them.

When opting for a BEV-based baseline model in the base branch, $F_{B}$ is generated as a BEV feature with the size of $(H_{B}, W_{B})$, and the feature fusion module integrates $F_{B}$ with object-aware voxel features $F_V(p)$.
Specifically, sparse 3D convolutions produce voxel features with valid values only in grids covered by pseudo-LiDAR points. Therefore, a significant number of invalid grids exist in $F_V$. To save memory, we use the valid values in $F_V(p)$ as queries and take the BEV features $F_B$ as values. Subsequently, we replace the corresponding values in $F_B$ with those calculated by voxel deformable self-attention (VSA) at $p=(x, y)$ to derive the BEV output $O_p$. The fusion process is as follows:
\begin{flalign}
    &&
        \mathcal{VSA}(F_V(p), F_B) = DA(F_V(p), p, F_B),
    &&
\end{flalign}
\begin{flalign}
    &&
        \begin{aligned}
O_p = \!
  \begin{cases}
    \mathcal{VSA}(F_V(p), F_B),\! &\text{if $F_V(p)$ is valid,}\\
	F_B(p),\! &\text{if $F_V(p)$ is invalid,}
  \end{cases}
\end{aligned}
    &&
\end{flalign}
where $DA$ denotes deformable attention.
Thus, foreground pseudo-3D features can be aggregated into the BEV representation through this approach.

When selecting a query-based baseline model, we sample flattened voxel features $F_V$ corresponding to the reference points of object queries at $p=(x, y)$, as shown in the following formula:
\begin{flalign}
    &&
        \mathcal{VSA}(F_I, F_V(p)) = DA(F_I, p, F_V(p)).
    &&
\end{flalign}

To take full advantage of the object-level depth estimation, we implement deformable spatial cross-attention~\cite{BEVFormer} on multi-camera depth maps. 
The depth features $F_D$ before softmax activation are employed as values, with $F_I$ or the flattened $F_B$ serving as the query $Q_p$. Each query generates $N_{ref}$ 3D sampling points, which are subsequently projected onto the corresponding depth feature maps to sample features from the respective views $\Gamma^{'}_h$.
A weighted sum is calculated for all depth feature points sampled by $Q_p$. The process of depth deformable cross-attention (DCA) can be expressed as:
\begin{flalign}
    &&
    \begin{aligned}
&\mathcal{DCA}(Q_p, F_D) = \\
&\frac{1}{\lvert\Gamma^{'}_{h}\rvert}\sum_{i\in \Gamma^{'}_{h}}\sum_{j=1}^{N_{ref}}\mathcal{DA}(Q_p, \mathcal{M}(p, i, j), F^i_D),
    \end{aligned}
    &&
\end{flalign}
where $\mathcal{M}$ is the camera's projection matrix.

\subsection{Loss Function}
Our loss function has three parts: (1) the 3D object detection loss $L_{3D}$, (2) the 2D object detection loss $L_{2D}$, and (3) the foreground depth classification loss $L_{dep}$. As such, it can be calculated as follows:
\begin{flalign}
    &&
        Loss = \alpha\cdot L_{3D} + \beta\cdot L_{2D} + \gamma\cdot L_{dep},
    &&
\end{flalign}
where $\alpha$, $\beta$, and $\gamma$ represent weight coefficients assigned to three distinct loss types. During training, these coefficients are adjusted to maintain a rough 20:4:1 ratio. The average ordinal loss of all foreground pixels in predicted depth distributions defines $L_{dep}$:
\begin{flalign}
    &&
    L_{dep} = \frac{1}{N} \sum_{n=0}^{N-1}\!
     \big(\sum_{k=0}^{l_{n-1}}log P^{k}_{(n,0)}\! +\! \sum_{k=l_{n}}^{K}log P^k_{(n,1)} \big),
    &&
\end{flalign}
where the product $N<H\times W$ denotes the total number of foreground pixels in the depth map, and $K$ is the number of discretized depth intervals.
\begin{table*}[t]
  \centering
    \caption{Comparison of different paradigms on the nuScenes \texttt{val} set. ST denotes Swin-Transformer~\cite{ST}. $\dag$ is initialized from a FCOS3D backbone. * notes that VoVNet-99 (V2-99)~\cite{V2-99} was pre-trained with external data~\cite{DD3D}. 
    $\ddag$ fine-tuned and tested with test time augmentation. The DETR3D, PETR, BEVDet, and OA-BEVDet are trained with CBGS~\cite{CBGS}. 
    ``mAP" and ``NDS" are reported in percentage (\%).
    }

    \setlength\tabcolsep{2.5pt}
    \begin{tabular}{l|cc|cc|ccccc}
    
    \toprule
    Methods &Image Size & Backbone  & \textbf{mAP}$\uparrow$ & \textbf{NDS}$\uparrow$ & mATE$\downarrow$  & mASE$\downarrow$   & mAOE$\downarrow$  & mAVE$\downarrow$  &  mAAE$\downarrow$  \\
    \midrule
    BEVDet-T\cite{BEVDet}       &704$\times$256  & ST-tiny   & 31.2  & 39.2    & 0.691             & 0.272              & 0.523             & 0.909             & 0.247       \\
    \textbf{OA-BEVDet-T (Ours)}      &704$\times$256    & ST-tiny  & \textbf{34.2} &  \textbf{41.1}         &    0.666          &     0.273    &    0.520     &   0.882      &     0.265            \\
    BEVDet$\dag$~\cite{BEVDet}       &1600$\times$640     & V2-99*  &  39.0 &  45.9  &  0.602    &    0.258  &0.382    & 0.901    &   0.216      \\
    \textbf{OA-BEVDet (Ours)}$\dag$     &1600$\times$640   & V2-99*  &   \textbf{40.2} &    \textbf{46.5}    &  0.597      &   0.252   &    0.355    &    0.932 &   0.220        \\
    \midrule
    BEVFormer-S$\dag$~\cite{BEVFormer}       &1280$\times$720  & R-101 &    38.9 &  49.2     &     0.700     &  0.275     &  0.390             &      0.424         &    0.196               \\   
    \textbf{OA-BEVFormer-S (Ours)}$\dag$     &1280$\times$720   & R-101 &  \textbf{40.7} & \textbf{51.0}               &    0.687               &      0.271       &    0.375        &       0.416        &     0.193             \\   
    BEVFormer$\dag$~\cite{BEVFormer}       &1600$\times$900   & R-101 & 41.6 & 51.7   & 0.673    & 0.274     &  0.372    & 0.394    & 0.198          \\

    \textbf{OA-BEVFormer (Ours)}$\dag$    &1600$\times$900 & R-101  & \textbf{43.1} &  \textbf{52.8}  &  0.664    &    0.272  &    0.388    & 0.345      &    0.205          \\
    \midrule
    DETR3D$\dag$~\cite{DETR3D}        &1600$\times$900    & R-101   & 34.9    & 43.4         & 0.716            & 0.268              & 0.379             & 0.842             & 0.200      \\
    PETR$\dag$~\cite{PETR}        &1600$\times$900  & R-101   & 37.0   & 44.2          & 0.711            & 0.267              & 0.383             & 0.865             & 0.201      \\
    Sparse4D~\cite{sparse4d} &1600$\times$900 & R101-DCN & 43.6 & 54.1 & 0.633 & 0.279 & 0.363 & 0.317 & 0.177 \\
    Sparse4Dv2~\cite{Sparse4Dv2} &704$\times$256    & R-50  & 43.9 & 53.9 & 0.598 & 0.270 & 0.475 & 0.282 & 0.179    \\ 
    
    StreamPETR-T~\cite{streampetr}  &704$\times$256    & R-50   & 44.9    & 54.6    & 0.618 & 0.274 & 0.420 & 0.267 & 0.208     \\
    \textbf{OA-StreamPETR-T (Ours)}  &704$\times$256    & R-50   & \textbf{45.5}    & \textbf{54.6}    & 0.618 & 0.269 & 0.469 & 0.261 & 0.198     \\
    StreamPETR-S~$\dag$\cite{streampetr}  & 800$\times$320  & V2-99   &  48.2  & 57.1  & 0.610 &  0.256  &  0.375 &  0.263  &   0.194  \\
    \textbf{OA-StreamPETR-S (Ours)}$\dag$  & 800$\times$320  & V2-99   &  \textbf{49.4}  & \textbf{57.6}  & 0.595 &  0.257  &  0.418 &  0.254  &   0.191   \\
    \midrule
    
    SparseBEV-T~\cite{sparsebev}  &704$\times$256    & R-50   & 44.8 & 55.8 & 0.581 & 0.271 & 0.373 & 0.247 & 0.190     \\  
    \textbf{OA-SparseBEV-T (Ours)}  &704$\times$256    & R-50   & \textbf{46.4} & \textbf{55.9}  & 0.597 & 0.267 & 0.424 & 0.249 & 0.191     \\  
    SparseBEV-S~\cite{sparsebev}  &  1408$\times$512   & R-101 & 50.1 & 59.2 & 0.562 & 0.265 & 0.321 & 0.243 & 0.195 \\  
    \textbf{OA-SparseBEV-S (Ours)}  &  1408$\times$512  & R-101   &  \textbf{50.8}  & \textbf{59.4}   &  0.558 &  0.260 & 0.355  &  0.236 &  0.190  \\  
    
    \bottomrule
    \end{tabular}
  \label{tab:nus-val}
  \vspace{-0.1cm}
\end{table*}

\section{Experiments}
\label{sec:experiments}
\begin{table*}[t]
  \centering
  \caption{Comparison with the state-of-the-art methods on the nuScenes \texttt{test} set. * notes that VoVNet-99 (V2-99)~\cite{V2-99} was pre-trained on the depth estimation task with extra data~\cite{DD3D}.
    $\ddag$ is test time augmentation.
    ``mAP" and ``NDS" are reported in percentage (\%).
    }
    \setlength\tabcolsep{2.5pt}
    \begin{tabular}{l|cc|cc|ccccc}
    \toprule
    Methods            &Image Size    & Backbone & \textbf{mAP}$\uparrow$ & \textbf{NDS}$\uparrow$   & mATE$\downarrow$   & mASE$\downarrow$  & mAOE$\downarrow$  & mAVE$\downarrow$  & mAAE$\downarrow$  \\
    
    \midrule
    
    BEVDet~\cite{BEVDet}             & 1600$\times$640 & ST-Small  &  39.8  & 46.3   & 0.556     & 0.239    & 0.414    & 1.010    & 0.153             \\

    BEVDet4D$\ddag$~\cite{BEVDet4D}   & 1600$\times$900  & ST-Base   & 45.1   & 56.9          & 0.511              & 0.241             & 0.386             & 0.301             & 0.121   \\   
    BEVFormer~\cite{BEVFormer}   & 1600$\times$900  & V2-99*   & 48.1   & 56.9          & 0.582              & 0.256             & 0.375             & 0.378             & 0.126                 \\
    \textbf{OA-BEVFormer (Ours)}   & 1600$\times$900 & V2-99*   &       \textbf{49.4}  &  \textbf{57.5}      &      0.574       &   0.256          &       0.377        &     0.385     &   0.132           \\
    
    \midrule
    
    DETR3D~\cite{DETR3D}   & 1600$\times$900   & V2-99*  & 41.2  & 47.9           & 0.641              & 0.255             & 0.394             & 0.845             & 0.133       \\
    PETR~\cite{PETR}   & 1600$\times$900   & V2-99*   & 44.1  & 50.4          & 0.593              & 0.249             & 0.383             & 0.808             & 0.132           \\
    MV2D~\cite{mv2d} & 1600$\times$640  & V2-99* & 46.3 & 51.4 & 0.542 & 0.247 & 0.403 & 0.857 & 0.127 \\
    UVTR~\cite{UVTR} & 1600$\times$900  & V2-99*   & 47.2   & 55.1 & 0.577 & 0.253 & 0.391 & 0.508 & 0.123 \\
    PETRv2~\cite{petrv2} & 1600$\times$900  & V2-99* & 49.0 & 58.2 & 0.561 & 0.243 & 0.361 & 0.343 & 0.120 \\
    PolarFormer~\cite{PolarFormer} & 1600$\times$900   & V2-99* & 49.3 & 57.2 & 0.556 & 0.256 & 0.364 & 0.439 & 0.127 \\
    Sparse4D~\cite{sparse4d} & 1600$\times$640   & V2-99* & 51.1 & 59.5 & 0.533 & 0.263 & 0.369 & 0.317 & 0.124 \\
    SparseBEV~\cite{sparsebev} & 1600$\times$640   & V2-99*  & 54.3 & 62.7 & 0.502 & 0.244 & 0.324 & 0.251 & 0.126 \\
    StreamPETR~\cite{streampetr} & 1600$\times$640   & V2-99*  & 55.0 & \textbf{63.6} & 0.479 & 0.239 & 0.317 & 0.241 & 0.119 \\
    \textbf{OA-StreamPETR (Ours)} & 1600$\times$640   & V2-99*  & \textbf{55.5}  & 63.1  & 0.489  &  0.242 &  0.349 &  0.269 & 0.116 \\
    \bottomrule
    \end{tabular}%
  \label{tab:nus-test}%
\end{table*}%

\begin{table*}[t]
  \centering
  \caption{Comparisons on the Argoverse 2 \texttt{val} set.  This study assesses 26 distinct object categories within a 150-meter detection range. Surround-view methods except for PETR are with temporal modeling.}

    \setlength\tabcolsep{6.1pt}
    \begin{tabular}{l|c|c|cc|ccc}
    \toprule
    Methods            &Image Size    & Backbone & \textbf{mAP(\%)}$\uparrow$ & \textbf{CDS(\%)}$\uparrow$   & mATE$\downarrow$   & mASE$\downarrow$  & mAOE$\downarrow$  \\
    
    \midrule
    
    BEVStereo~\cite{BEVStereo}   & 960$\times$640 & VoV-99  &  14.6  &  10.4   & 0.847  &  0.397  & 
 0.901            \\
    SOLOFusion~\cite{SOLOFusion}   & 960$\times$640 & VoV-99  & 14.9 & 10.6 & 0.934 & 0.425 & 0.779                 \\

    PETR~\cite{PETR}   & 960$\times$640 & VoV-99   & 17.6 & 12.2 & 0.911 & 0.339 & 0.819 \\

    Sparse4Dv2~\cite{Sparse4Dv2}   & 960$\times$640 & VoV-99   & 18.9 & 13.4 & 0.832 & 0.343 & 0.723   \\   
    StreamPETR~\cite{streampetr} & 960$\times$640 & VoV-99   & 20.3 & 14.6 & 0.843 & 0.321 & 0.650    \\

    \textbf{OA-StreamPETR (Ours)}   & 960$\times$640 & VoV-99  & \textbf{21.5} & \textbf{15.7} & 0.815 & 0.322 & 0.602     \\
    
    \bottomrule
    \end{tabular}
  \label{tab:av2}
\end{table*}

\subsection{Datasets and Metrics}
\parsection{nuScenes} 
We perform experiments on the large-scale autonomous driving dataset, nuScenes~\cite{nuScenes}, which encompasses various perception tasks, including detection, tracking, and LiDAR segmentation. The nuScenes dataset comprises 1,000 scenes, separated into training (700), validation (150), and testing (150) sets. Each scene consists of 20 seconds of perceptual data annotated with a keyframe at 2 Hz. The data acquisition vehicle features one LiDAR, six cameras that form a surround view, and five radars. To assess detection errors, nuScenes proposes several true positive metrics, including ATE, ASE, AOE, AVE, and AAE, to measure translation, scale, orientation, velocity, and attribute errors, respectively. Furthermore, the mean average precision (mAP) and nuScenes detection score (NDS) are comprehensive metrics utilized to evaluate detection performance, where higher scores indicate superior performance.

\parsection{Argoverse 2}
Argoverse 2~\cite{av2} is a comprehensive benchmark tailored to advance research on autonomous vehicle perception and motion prediction. It comprises three distinct yet interconnected datasets: the annotated sensor dataset, the Lidar dataset, and the motion forecasting dataset. The sensor dataset features 1,000 multi-modal sequences, each captured through an array of seven high-resolution ring cameras, two stereo cameras, and a LiDAR system, all aligned with a 6-DOF map. These sequences are annotated with 3D bounding box labels in 26 object categories, facilitating robust 3D perception model training and validation. Argoverse 2 is structured into 700 training, 150 validation, and 150 testing scenes, each spanning 15 seconds at a 10Hz annotation frequency. Evaluation metrics include the composite detection score (CDS) and three true positive metrics--ATE, ASE, and AOE--complementing the traditional mAP.

\subsection{Implementation Details}
The experiments on nuScenes employ a BEV perception range of $[-51.2m, 51.2m]$ for $X$ and $Y$ axes and $[-5m, 3m]$ for $Z$ axis. We evaluate the average detection results across ten categories, including car, truck, bicycle, and pedestrian, among others.
The BEV perception range set for Argoverse 2 spans from -151.4 meters to 151.4 meters on both the $X$ and $Y$ axes, and from -5 meters to 5 meters on the $Z$ axis. The dataset features a detailed classification into 26 distinct categories, encompassing relatively rare objects in autonomous driving contexts such as dogs, wheelchairs, and strollers.

\parsection{Implementation Details of BEV-based Baseline Models}
To assess the impact of object-aware features, we implement \systemname on two BEV-based baseline models, namely BEVDet~\cite{BEVDet} and BEVFormer~\cite{BEVFormer}. Our models are named \textit{OA-BEVDet} and \textit{OA-BEVFormer}, respectively, based on the above networks.

For its tiny and base versions, BEVDet employs SwinTransformer-Tiny~\cite{ST} and ResNet101~\cite{ResNet}, respectively, as backbones, followed by an FPN-LSS~\cite{BEVDet}. The input resolutions for the two versions are $704\times256$ and $1600\times640$, respectively. This BEV-based pipeline implements lifting-collapsing as the view transformation method, a combination of a lightweight ResNet and an FPN as the BEV encoder, and center point head~\cite{CenterPoint3D} as the BEV detection head. BEVDet divides the ground plane with a resolution of 0.4 meters for the base version and 0.8 meters for the tiny version during view transformation.

BEVFormer-Small adopts ResNet101-DCN~\cite{ResNet, DCN} as the backbone, while experiments are conducted with VoVNet-99 (V2-99)~\cite{V2-99} as the backbone for the base version. Input resolutions are set to $1280\times720$ and $1600\times900$ for the two versions, respectively. To perform view transformation, stacked spatial cross-attention layers are utilized, and the DETR head~\cite{DETR} is used as the BEV detection head. For view transformation, BEVFormer-Base sets the size of BEV queries to $200\times200$. To maintain the same size as the output of the 3D voxel network in \systemname, we set the size of BEV queries to $160\times160$ for BEVFormer-Small. Additionally, BEVFormer aggregates temporal information to recurrently fuse historical BEV features.

\parsection{Implementation Details of Query-based Baseline Models}
We integrate \systemname into two existing query-based models: SparseBEV~\cite{sparsebev} and StreamPETR~\cite{streampetr}. The resulting models, termed \textit{OA-SparseBEV} and \textit{OA-StreamPETR}, respectively, are enhanced versions of the original networks incorporating our general plug-in modules.

We perform experiments using various backbone networks, including ResNet50, ResNet101, and V2-99~\cite{vovnet}, each under respective pre-training conditions. 
Consistent with prior studies, the ResNet50 and ResNet101 models are pre-trained with nuImages~\cite{nuScenes} weights. 
Additionally, to demonstrate the scalability of our approach, we extend our evaluation to the nuScenes \texttt{test} set, utilizing the V2-99 backbone initialized from the DD3D~\cite{DD3D} checkpoint.
For Argoverse 2, we use the V2-99 weights pre-trained in DDAD~\cite{DDAD} and further trained on the nuScenes train set with FCOS3D~\cite{FCOS3D}, consistent with previous works~\cite{PETR, streampetr}.

\parsection{Implementation Details of \systemname} 
We set up three versions of \systemname: tiny, small, and base, each with varying input resolutions, backbones, feature dimensions, and sizes of BEV grids.

\systemname incorporates several key modules into its general pipeline, including a 2D objection head, a depth head with object-level supervision, an object-centric pseudo-3D feature generator, and an object-aware feature fusion module. To identify foreground regions, we introduce an FPN with 256 channels and an FCOS head after the backbone to predict 2D bounding boxes. The depth head is lightweight and comprises only one convolutional layer with $K\times2$ channels, followed by a softmax operation. Here, $K=80$ denotes the number of depth bins in our network. In the object-centric pseudo-3D feature generator, we select foreground pixels in the depth maps using predicted 2D bounding boxes to generate foreground 3D pseudo-LiDAR point clouds. Subsequently, we voxelize the pseudo point clouds before feeding them to a voxel network composed of four layers of 3D sparse convolutions to obtain an $8\times$ down-sampled volume. We flatten the volume along the z-axis to produce a BEV output. For the object-aware feature fusion module, we use three encoder layers, each following the conventional structures of deformable transformer~\cite{Deformable-DETR}. Attention layers are replaced with voxel deformable self-attention and depth deformable cross-attention.

\begin{table}[t]
  \centering
  \caption{
  Ablation analysis of the crucial modules of \systemname on the nuScenes \texttt{val} dataset. The acronyms ODS, DF, and PF respectively denote object-level depth supervision, depth feature fusion, and pseudo-3D feature fusion.
  }
    \setlength{\tabcolsep}{12.5pt}{
    \begin{tabular}{ccc|cc}
    \toprule
    ODS       & DF  & PF   & mAP(\%) & NDS(\%) \\
    \midrule
         &     &      & 31.2 & 39.2      \\
    \checkmark    &      &  & 32.6  &  39.8\\
    \checkmark   & & \checkmark &   33.8  &  40.7 \\
    \checkmark  &     \checkmark      &     \checkmark  & \textbf{34.2} &\textbf{41.1}     \\
    \bottomrule
    \end{tabular}
    }%
  \label{tab:ablation}
\end{table}

\parsection{Training}
We broadly follow the training strategies of the four baselines. All models are trained with AdamW~\cite{adamw} optimizer and a learning rate of 2e-4. The learning rate multiplier for the backbones V2-99 and ResNet-101 in the base version is 0.1. When using BEVDet and StreamPETR as baselines, we train the base models with a batch size of 1 per GPU on 8 NVIDIA GeForce RTX 3090 GPUs. When using BEVFormer and SparseBEV as baselines, we decay the learning rate using cosine annealing~\cite{cosinedecay}, with a batch size of 1 per GPU on 8 NVIDIA GeForce RTX A6000 GPUs. We perform end-to-end training across the entire model.

\begin{figure}[t]
    \centering
    \includegraphics[width=0.95\columnwidth]{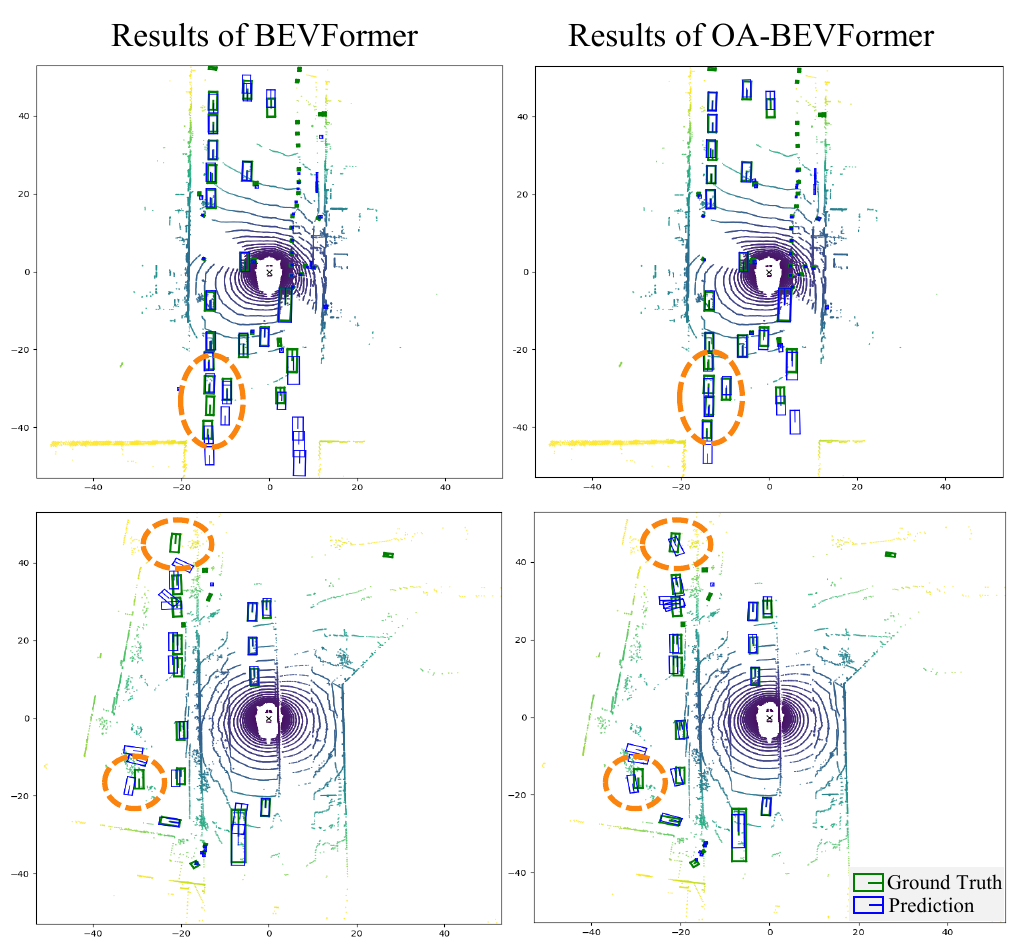}
    \caption{Qualitative analysis of the detection results from OA-BEVFormer and BEVFormer in the small version.} 
    \label{fig:vis}
\end{figure}

\subsection{Main Results}

\parsection{Comparison on nuScenes Dataset}
We conduct comparisons of \systemname with current state-of-the-art multi-camera 3D perception methods, utilizing the nuScenes \texttt{val} and \texttt{test} sets. All methods used multi-camera images as input and ground-truth 3D objects for supervision during training.

Let us first look at the results of the \texttt{val} set shown in Table~\ref{tab:nus-val}. Here, we conduct four groups of comparison experiments since we evaluate our plug-in module on four baselines.
The first group compares the performance between OA-BEVDet and BEVDet. 
All methods in this group perform multi-camera 3D object detection without incorporating time-series data fusion.
The networks of the tiny version share the same backbone, SwinTransformer-Tiny, with a low input resolution of $704\times256$. Additionally, the small size of the BEV features at $128\times128$ enhances efficiency but reduces the model's ability to describe objects due to the coarse granularity. Our results show that OA-BEVDet-Tiny outperforms BEVDet significantly in terms of mAP and NDS, by 3.0\% and 1.9\%, respectively.
As shown in the last two rows of the first group, we reproduce the performance of BEVDet-Base utilizing V2-99 as the backbone and use this as our baseline to evaluate the effect of object-aware features. Our results show that OA-BEVDet achieved improvements of 1.2\% and 0.6\%  in mAP and NDS, respectively.

BEVFormer leverages temporal information, which greatly enhances the network's performance compared to BEVDet, particularly in terms of lower mAVE. In the second group, we compare the small and base versions of OA-BEVFormer against BEVFormer, respectively. Without modifying the use of temporal information, OA-BEVFormer-Small outperforms BEVFormer-Small in both mAP and NDS by 1.8\%, while OA-BEVFormer achieves a similar improvement of 1.5\% and 1.1\% for the base version.

The third and fourth groups present a comparison between query-based methods. In the tiny version, OA-SparseBEV outperforms SparseBEV with an mAP increase of 1.6\%, while OA-StreamPETR shows a 1.2\% higher mAP than StreamPETR. In the small version, the improvement is 0.6\% for OA-SparseBEV over SparseBEV and 0.7\% for OA-StreamPETR over StreamPETR in terms of mAP. Synthesizing the results from all groups, it is evident that our method yields more significant gains for the tiny version with a small resolution input and a lightweight backbone.

\begin{figure}[t]
    \centering
    \includegraphics[width=.97\columnwidth]{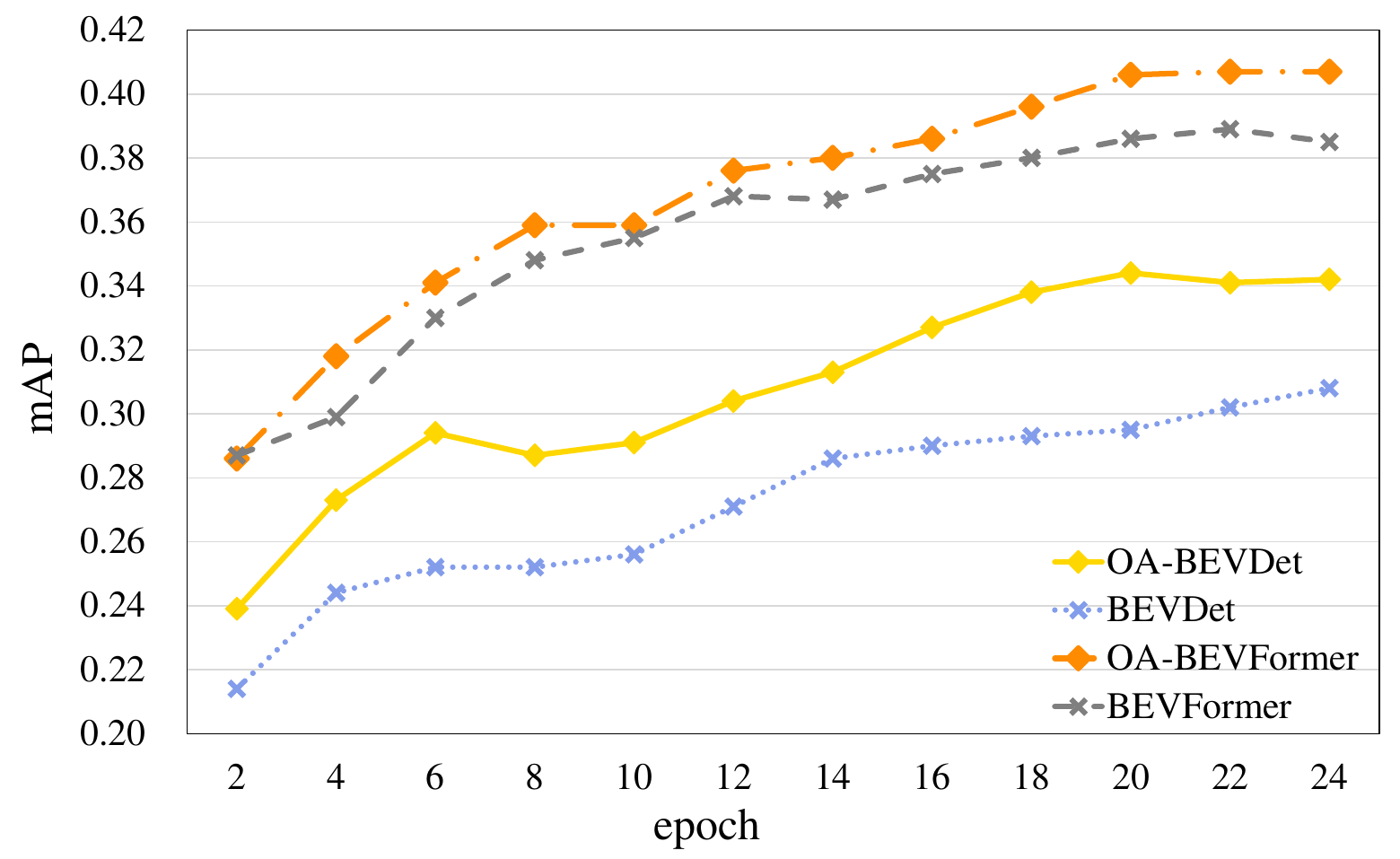}
    \caption{The evolution of mAP in the training process. We compare the intermediate results in the lightweight versions of \systemname and its two baseline networks.}
    \label{fig:intermediate_results}
    \vspace{-0.1cm}
\end{figure}

Figure~\ref{fig:intermediate_results} illustrates the changes in mAP during the training process for \systemname based on BEVDet-Tiny and BEVFormer-Small. Our results demonstrate that our method exhibits advantages early during training and converges faster compared to the baselines.

Table~\ref{tab:nus-test} summarizes the \texttt{test} set results.
To generate these results, we use BEVFormer as a baseline for the BEV-based methods in the first group and StreamPETR as a baseline for query-based methods in the second group. The models are trained on both the \texttt{train} and \texttt{val} sets. Without any test time augmentation or model ensemble, our OA-BEVFormer outperforms BEVFormer with V2-99 by 1.3\% in mAP and 0.6\% in NDS, respectively. 
Compared with query-based methods in the second group, OA-StreamPETR has achieved 55.5\% in mAP and 63.1\% in NDS, surpassing the baseline model StreamPETR by 0.5\% in mAP.
Based on these findings, we conclude that \systemname consistently enhances detection performance in terms of mAP and NDS when combined with various base models across different settings.

\begin{table*}[t]
  \centering
  \caption{Comparisons of \textbf{mAP} and \textbf{NDS} across different weather and lighting conditions on the nuScenes \textit{val} set.
  ``mAP" and ``NDS" are reported in percentage (\%).
  }

    \setlength{\tabcolsep}{14.7pt}{
    \begin{tabular}{l|cc|cc|cc|cc}
    \toprule
    \multirow{2}{*}{Methods} & \multicolumn{2}{c|}{Night} & \multicolumn{2}{c|}{Day} & \multicolumn{2}{c|}{Rainy} &  \multicolumn{2}{c}{Sunny}\\
    \cline{2-9}
    & mAP & NDS & mAP & NDS & mAP& NDS & mAP & NDS\\
    \midrule
    BEVDet & 13.4 & 23.2 & 31.2 & 40.7 & 32.4   & 46.4 & 30.1  & 39.2  \\
    OA-BEVDet & \textbf{15.6} & \textbf{24.4} & \textbf{32.1} & \textbf{41.5} & \textbf{35.5}   & \textbf{48.1} & \textbf{30.9}  & \textbf{40.2}  \\
    StreamPETR & 29.2 & 35.1 & 48.5 & 57.5 & 55.2   & 62.4 & 46.8  & 56.0  \\
    OA-StreamPETR & \textbf{30.3}  & \textbf{36.1} & \textbf{49.7} & \textbf{57.9} & \textbf{57.0} & \textbf{63.5} & \textbf{48.1} & \textbf{56.4} \\
    SparseBEV & 25.5  & 34.1  &  45.9 &  55.9 & 51.1 &  60.2 &  44.2 & 54.6  \\
    OA-SparseBEV & \textbf{26.9} & \textbf{35.5} & \textbf{46.8} & \textbf{56.2} & \textbf{52.6} & \textbf{62.1} & \textbf{45.2} & \textbf{54.6} \\
    \bottomrule
    \end{tabular}
    }
  \label{tab:condi}
\end{table*}

\begin{table}[t]
  \centering
    \caption{Comparisons of different distances on the nuScenes \texttt{val} set.
    ``mAP" and ``NDS" are reported in percentage (\%).
    }
    \setlength\tabcolsep{3.6pt}
    \begin{tabular}{l|cc|cc|cc}
    \toprule
    \multirow{2}{*}{Methods} & \multicolumn{2}{c|}{0-15m} & \multicolumn{2}{c|}{15-30m} & \multicolumn{2}{c}{30-50m}\\
    \cline{2-7}
    & mAP & NDS & mAP & NDS & mAP& NDS \\
    \midrule
    BEVDet & 53.6 & 50.8 & 28.5 & 39.6 & 6.1 & 19.0 \\
    OA-BEVDet & \textbf{56.4} & \textbf{52.9} & \textbf{31.9}  & \textbf{40.2} & \textbf{7.1}  & \textbf{20.9}  \\
    BEVFormer & 56.4 & 58.7 & 36.4 & 48.0 & 10.2  & 27.0  \\
    OA-BEVFormer & \textbf{57.5} & \textbf{61.1} & \textbf{38.8} & \textbf{50.3}& \textbf{11.6} & \textbf{27.8} \\
    \midrule
    StreamPETR & 64.9 & \textbf{67.6} & 45.4 & 55.5 & 12.7   & 31.4  \\
    OA-StreamPETR & \textbf{65.4}  & 66.9  & \textbf{46.7} & \textbf{56.8} & \textbf{14.2}  & \textbf{33.8} \\
    SparseBEV & 65.4 & 67.0 & 43.1 & 54.8 &  17.3  & 34.3   \\
    OA-SparseBEV & \textbf{65.8}  & \textbf{67.6}  & \textbf{43.4} & \textbf{54.8} &  \textbf{19.3} & \textbf{37.0} \\
    \bottomrule
    \end{tabular}
  \label{tab:distance}
\vspace{-0.2cm}
\end{table}

\parsection{Comparison on Argoverse 2 Dataset}
In Table~\ref{tab:av2}, we compare our proposed \systemname with other state-of-the-art methods on the Argoverse 2 dataset\footnote{The results of BEVStereo and SOLOFusion are reproduced by Far3D~\cite{far3d}.}. When equipped with a VoV-99 backbone and an input resolution of $960\times640$, \systemname achieves absolutely 1.2\% improvement in mAP and 1.1\% improvement in CDS compared to the strongest baseline, \emph{i.e.}, StreamPETR~\cite{streampetr}.
Furthermore, it is observable that OA-StreamPETR exhibits a notable decrease in both mATE (mean average translation error) and mAOE (mean average orientation error). 
Given that the Argoverse 2 dataset requires a larger detection range compared to the nuScenes dataset, it introduces greater challenges in feature representation due to increased distortion at farther distances. By enhancing object awareness and refining depth estimation at the object level, our approach improves precision in object localization and orientation accuracy.

\subsection{Visualization Results}
Figure~\ref{fig:vis} shows some qualitative detection results of OA-BEVFormer and BEVFormer in BEV. Our analysis indicates that OA-BEVFormer eliminates some of the missed detections (circled in orange) of distant objects compared to BEVFormer, as demonstrated in the visualizations.
The visualization results in Figure~\ref{fig:vis_more} demonstrate the detection performance of OA-BEVFormer in various scenarios, including rainy scenes, complex scenes, scenes with many small objects such as pedestrians and traffic cones, and scenes with truncated objects.
Facing challenging weather conditions, OA-BEVFormer proves its resilience by accurately detecting and localizing objects within rainy scenes. In the context of complex scenes with multiple objects and occlusions, the network exhibits its proficiency in accurately identifying and classifying objects.
Moreover, OA-BEVFormer continues to show commendable detection and localization performance in scenes that contain truncated objects. Inevitably, it encounters difficulties when dealing with scenes abundant with small or distant objects, indicating an area where the network's performance could be further improved.

\begin{table}[t]
  \centering
  \caption{Comparisons of two object-level depth supervision methods on the nuScenes \texttt{val} set.
  ``mAP" and ``NDS" are reported in percentage (\%).
  }

    \setlength{\tabcolsep}{2.6pt}{
    \begin{tabular}{l|l|cc}
    \toprule
    Models & Methods & mAP& NDS \\
    \midrule
    \multirow{2}{*}{OA-BEVDet} & Center Point Diffusion & 34.2 & 41.1 \\
    & Diagonal Multi-Point Slicing  & 34.5 & 41.1 \\
    \midrule
    \multirow{2}{*}{OA-BEVFormer} & Center Point Diffusion& 40.7 & 51.0\\
    & Diagonal Multi-Point Slicing  & 40.8 & 51.2 \\        
    \bottomrule
    \end{tabular}
    }
  \label{tab:depth_supervision}
\end{table}

\begin{table}[t]
  \centering
  \caption{Ablation study of the object-aware feature fusion module.}
    \setlength{\tabcolsep}{14pt}{
    \begin{tabular}{l|cc}
    \toprule
    Fusion Methods & mAP(\%) & NDS(\%) \\
    \midrule
    Concatenation & 33.0  & 39.2  \\
    Addition  & 33.4 & 39.9 \\
    Deform. Transformer & 34.2 & 41.1 \\
    \bottomrule
    \end{tabular}
    }
  \label{tab:fusion_module}
\end{table}

\subsection{Ablation Studies}
To quantify the impact of each module on the performance of \systemname, we summarize the results of our ablation studies in Table~\ref{tab:ablation}.

\begin{figure*}[t]
    \centering
    \setlength{\tabcolsep}{4pt}
    \begin{tabular}{cc}
    \includegraphics[width=0.95\columnwidth]{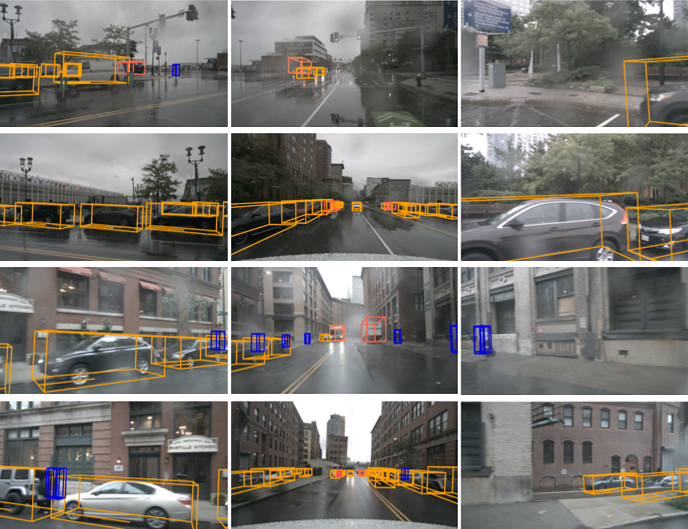} &
    \includegraphics[width=0.95\columnwidth]{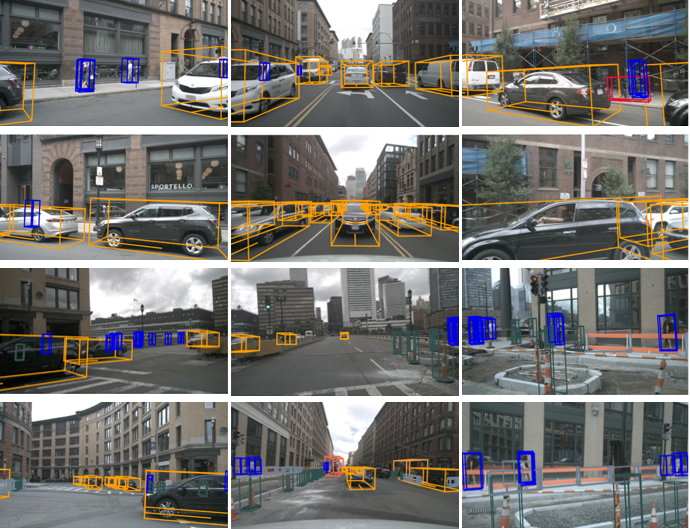} \\
    (a) Rainy scenes & (b) Complex scenes \\
    \includegraphics[width=0.95\columnwidth]{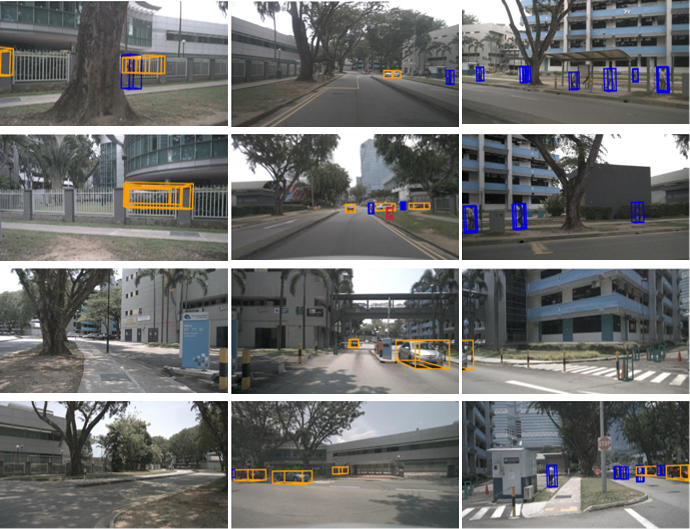} & 
    \includegraphics[width=0.95\columnwidth]{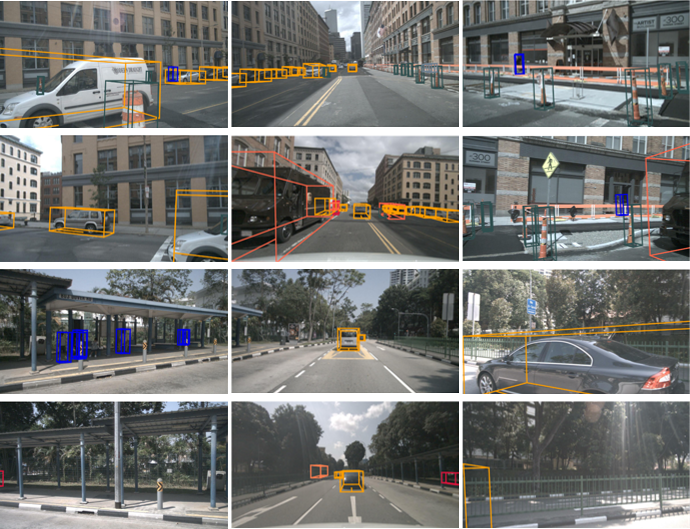}\\
    (c) Scenes with many small objects & (d) Scenes with truncated objects \\
    \end{tabular}
    \caption{
    Qualitative analysis of OA-BEVFormer's detection results for different scenarios: rainy scenes, complex scenes, scenes with many small objects, and scenes with truncated objects.
    } 
    \label{fig:vis_more}
\end{figure*}

\parsection{Object-Level Depth Supervision} 
We compare the results obtained by our network with object-level depth supervision against those produced by a baseline model without such supervision. Our experiments, reported in Table~\ref{tab:ablation}, reveal that the inclusion of object-level depth supervision leads to improvements of 1.4\% and 0.6\% in mAP and NDS metrics, respectively. 
Table~\ref{tab:depth_supervision} provides a comparative analysis of the efficacy of two supervision methods: center point diffusion and diagonal multi-point slicing, as applied to the OA-BEVDet and OA-BEVFormer models. The findings suggest that both methods generate comparable performance outcomes.
However, the diagonal multi-point slicing method, which selects $N=16$ points along the diagonal, exhibits a slight edge due to its finer granularity. Despite this, we have chosen to employ the center point diffusion method as the default technique for the remainder of this paper, due to its balance between performance and complexity.

\parsection{Integration of Pseudo-3D Features}  
We employ voxel deformable self-attention to merge the pseudo-3D features. 
Our experimental findings, as presented in Table~\ref{tab:ablation}, demonstrate a significant enhancement in detection accuracy by incorporating foreground pseudo-3D features, particularly in terms of mAP (+1.2\%). This observation supports the notion that pseudo-3D features have a positive influence on object identification by the network.

\parsection{Integration of Depth Features}
The depth features are integrated with BEV features by means of depth deformable cross-attention. Comparison of the penultimate and final rows of Table~\ref{tab:ablation} delineates the impact of depth feature integration, demonstrating a positive effect on both mAP and NDS by 0.4\%. These findings signify that the depth distribution may be utilized not only for view transformation but also can be directly encoded into the BEV representation to enhance performance even further.

\parsection{Object-Aware Feature Fusion Module}
We present a comparative analysis of three distinct fusion methods: concatenation, addition, and deformable attention. Our findings, as illustrated in Table~\ref{tab:fusion_module}, reveal that the employment of the deformable transformer yields $1.2\%$ and $0.8\%$ improvement in mAP compared to concatenation and addition methods, respectively. Moreover, it manifests $1.9\%$ and $1.2\%$ performance gains in NDS over the aforementioned techniques, respectively. These results underscore the efficacy of the deformable transformer in enhancing the performance of the fusion process.

\begin{table}[t]
  \centering
  \caption{Ablation study of the foreground region selection using 2D object detection.}
    \setlength{\tabcolsep}{5.8pt}{
    \begin{tabular}{l|l|cc}
    \toprule
    ID& Methods & mAP(\%) & NDS(\%) \\
    \midrule
    A&OA-BEVDet (ours) & 34.2 & 41.1 \\
    B&OA-BEVDet w/o 2D Det. & 31.5 & 38.9 \\
    C&OA-BEVDet w/ 2D GT Box & 36.2 & 41.4 \\
    \bottomrule
    \end{tabular}
    }
  \label{tab:2d_abla}
\end{table}

\begin{table}[t]
  \centering
  \caption{Ablation study on input resolution. The results are evaluated on the nuScenes \texttt{val} set. We use ResNet-50 as the backbone network by default. The resolution of BEV feature is $128\times128$.
  }
    \setlength{\tabcolsep}{10pt}{
    \begin{tabular}{c|c|cc}
    \toprule
    Inp. Res. & Methods & mAP(\%) & NDS(\%) \\
    \midrule
    \multirow{2}{*}{704$\times$256} & BEVDet & 30.2 & 39.0\\
      & OA-BEVDet & 34.2 & 41.1\\
    \midrule
     \multirow{2}{*}{1056$\times$384} & BEVDet & 33.3 & 41.0 \\
      & OA-BEVDet & 36.0 & 41.8\\
    \bottomrule
    \end{tabular}
    }
  \label{tab:input_resolution}
\end{table}

\parsection{Different Weather and Light Conditions}
To verify the robustness of \systemname across various scenarios, we categorize the nuScenes validation set into four scenarios based on weather and lighting conditions: night, day, rainy, and sunny. We evaluate \systemname using three baselines—BEVDet, StreamPETR, and SparseBEV—on the nuScenes validation set. In Table~\ref{tab:condi}, \systemname demonstrates improvements over the baselines in all scenarios, particularly in night and rainy conditions. Specifically, in the rainy scenario, mAP increased by 3.1\%, 1.8\%, and 1.5\% for OA-BEVDet, OA-StreamPETR, and OA-SparseBEV, respectively. In the night scenario, the improvements were 2.2\%, 1.1\%, and 1.4\% respectively.

\parsection{Different Depth Ranges}
Table~\ref{tab:distance} presents the mAP improvements for four distinct methods, encompassing both BEV-based and query-based approaches, across various depth intervals. Specifically, for BEV-based methods, OA-BEVDet shows relative mAP gains of 5.2\%, 11.9\%, and 16.4\% in the depth ranges of 0-15m, 15-30m, and 30-50m, respectively. On the other hand, OA-BEVFormer's relative mAP gains in these ranges are 1.2\%, 5.1\%, and 5.5\%, respectively.
In the case of query-based methods, OA-StreamPETR achieves relative mAP gains of 0.8\%, 2.9\%, and 11.8\% across the same depth ranges. Similarly, OA-SparseBEV records relative mAP gains of 0.6\%, 0.7\%, and 11.6\% for the 0-15m, 15-30m, and 30-50m depth ranges, respectively.
The results demonstrate that our system significantly enhances the performance of both BEV-based and query-based detection methods across the full range of detection distances.
Notably, \systemname exhibits a more pronounced enhancement in detection accuracy at greater distances when compared to the baseline models. This highlights the particular advance of \systemname in scenarios that demand reliable detection over extended ranges.

\parsection{Foreground Region Selection in Images} 
We train three models for the comparative experiments: (A) OA-BEVDet as the baseline, (B) OA-BEVDet without 2D object detection, and (C) OA-BEVDet using 2D ground-truth bounding boxes. The results in Table~\ref{tab:2d_abla} reveal that our proposed OA-BEVDet model outperforms model B by 2.7\% mAP and 2.2\% NDS. Moreover, model C yields superior outcomes compared to model A, with an improvement of 2.0\% mAP and 0.3\% NDS. The significant variations in mAP observed in these experiments validate the criticality of our object-aware paradigm for accurate object identification.

\begin{table}[t]
  \centering
  \caption{Ablation study on BEV feature resolution. The results are evaluated on the nuScenes \texttt{val} set. We use ResNet-50 as the backbone network with the input resolution of $704\times256$ by default.}
    \setlength{\tabcolsep}{10pt}{
    \begin{tabular}{c|c|cc}
    \toprule
    BEV Res. & Methods & mAP(\%) & NDS(\%) \\
    \midrule
     \multirow{2}{*}{128$\times$128} & BEVDet & 30.2 & 39.0 \\
      & OA-BEVDet & 34.2 & 41.1 \\
    \midrule
     \multirow{2}{*}{160$\times$160} & BEVDet & 30.7 & 38.3 \\
     & OA-BEVDet & 34.7 & 41.7 \\
    \midrule
     \multirow{2}{*}{200$\times$200} & BEVDet & 31.3 & 38.7 \\
      & OA-BEVDet & 35.0 & 41.5 \\
    \bottomrule
    \end{tabular}
    }
  \label{tab:bev_resolution}
  \vspace{-0.2cm}
\end{table}

\parsection{Input Resolution}
We complement the comparative experiments at resolution $1056\times384$ on the nuScenes \texttt{val} set. In Table~\ref{tab:input_resolution}, the performance of OA-BEVDet was compared to the baseline network BEVDet by changing the input resolution of multi-view images without altering network structure and settings. At $704\times256$ resolution, OA-BEVDet improves mAP and NDS by 4.0\% and 2.1\%, respectively, while at $1056\times384$ resolution, it outperforms BEVDet by 2.7\% and 0.8\%, respectively. The performance improvement of \systemname is more significant at low input resolutions.

\parsection{BEV Resolution}
Table~\ref{tab:bev_resolution} shows the impact of different BEV resolutions on detection performance, comparing it with the BEV-based baseline network BEVDet at the same resolution. Both the baseline and OA-BEVDet show improved mAP with increased BEV resolution, while changes in NDS show little regularity. This is consistent with the conclusion that object awareness mainly affects object recognition and positioning. At a BEV resolution of $160\times160$, OA-BEVDet achieves an mAP of 34.7\%. This performance is 0.5\% superior compared to the detection results obtained at a lower BEV resolution of $128\times128$. Furthermore, at a BEV resolution of $200\times200$, OA-BEVDet achieves an mAP of 35.0\%, which is 0.3\% higher than at a BEV resolution of $128\times128$. These results highlight the \systemname's enhanced detection accuracy for the BEV-based methods with the increase in BEV resolution.

\subsection{Runtime details}
We undertake a comparative analysis between our proposed methods and their respective base versions, and the findings pertaining to Frames Per Second (FPS) and the number of parameters (M) are presented in Table~\ref{tab:latency}. The FPS for OA-BEVDet and OA-BEVFormer are measured using NVIDIA GeForce RTX 3090 GPU and NVIDIA RTX A6000 GPU, respectively.
As our method serves as a plug-in module integrated with other methods to enhance object awareness, an increase in the runtime latency and the number of parameters is to be expected.
On a general note, the incorporation of \systemname results in an approximate latency increase of 20\%. 
Furthermore, OA-BEVDet has 71.3M parameters, which is a 6.9M increase compared to BEVDet, while OA-BEVFormer has 77.8M parameters, marking a 9.1M increase compared to BEVFormer.
The increase in parameters for both OA-BEVDet and OA-BEVFormer is primarily attributed to the additional network components required for estimating object-level depth and generating foreground pseudo-3D features. 
These components are deliberately designed to enhance the model's generalizability and robustness. Our method's ability to be loosely coupled with various baseline models as a plug-in module exemplifies its effectiveness.
We are committed to ongoing optimization to ensure that the benefits of our method are realized without undue burden on computational resources.

\begin{table}[t]
  \centering 
  \caption{The FPS and parameters (M) of models in base versions.
  ``mAP" is reported in percentage (\%).
  }
    \setlength{\tabcolsep}{1.8pt}
    \begin{tabular}{l|cc|ccc}
    \toprule
    Models & Input & Backbone & mAP & FPS & \#param. \\
    \midrule
    BEVDet & 1600$\times$640 & R-101 & 39.0 & 4.5 & 64.4 \\
    OA-BEVDet  & 1600$\times$640 & R-101 & 40.2 & 3.7 & 71.3\\
    BEVFormer & 1600$\times$900 & V2-99 & 41.6 & 2.6  & 68.7\\
    OA-BEVFormer & 1600$\times$900 & V2-99 & 43.1 & 2.2 & 77.8\\
    \bottomrule
    \end{tabular}
  \label{tab:latency}
\end{table}

\begin{table}[t]
  \centering
  \small
    \caption{
    Results of OA-BEVDet using external depth models including Depth Anything and UniDepth V2. ``mAP" and ``NDS" are reported in percentage (\%).
    }

    \setlength\tabcolsep{2pt}
    \begin{tabular}{c|ccccc}
    
    \toprule
     External Dep. & \textbf{mAP}$\uparrow$ & \textbf{NDS}$\uparrow$ & mATE$\downarrow$     & mAOE$\downarrow$  & mAVE$\downarrow$  \\
    \midrule
    -  &   34.2  &   41.1   &    0.666  &  0.520     &   0.882   \\

     Depth Anything  & 33.5   &   40.6  &    0.682      & 0.510 &    0.919  \\
     UniDepth V2 &  33.7   &   40.7   &    0.678      & 0.503 &    0.895   \\

    \bottomrule
    \end{tabular}
  \label{tab:foundation_0}
\end{table}

\subsection{External Depth Models}

In this section, we study the effect of depth foundation models and present the results in Table~\ref{tab:foundation_0}.
Specifically, we \textbf{replace the lightweight depth estimation module in our pipeline with either Depth Anything or UniDepth V2} and finetune the whole framework on nuScenes.

When replacing our lightweight depth module with general-purpose foundation models such as Depth Anything or UniDepth V2, we observe a performance drop, with mAP reductions of 0.7\% and 0.5\%, respectively. This degradation can be primarily attributed to two factors. First, our object-level supervision relaxes the requirement for pixel-accurate depth estimation--especially along instance boundaries--which reduces the optimization benefits that foundation models typically provide. More importantly, removing our original depth estimation head eliminates the model’s ability to retain and utilize depth-specific inductive biases learned during task-oriented training. As a result, both semantic understanding and object localization suffer.

In summary, our method offers two key advantages. First, we introduce an object-aware plug-in module that only relies on 3D instance-level annotations. This design delivers strong detection performance without relying on the precise and expensive depth annotations typically needed by foundation models like Depth Anything or UniDepth V2. Second, by jointly modeling depth estimation, 2D object detection, and 3D object detection within the object-aware branches, our framework makes the best of the complementary nature of these tasks. The shared multi-task supervision allows the model to learn richer and more task-relevant representations, ultimately improving 3D localization accuracy. These results further highlight the effectiveness of our task-driven training strategy and the benefit of jointly optimizing depth and detection for robust 3D perception.

\begin{table}[t]
  \centering
    \caption{
    Ablation study on the impact of lightweight optimization and image-view predictive temporal enhancement (IPTE) on accuracy and inference speed (FPS). ``mAP" and ``NDS" are reported in percentage (\%).
    }
    \setlength\tabcolsep{2.2pt}
    \begin{tabular}{l|cc|cccc}
    \toprule
    Methods & \textbf{mAP} & \textbf{NDS} & mATE & mAOE & mAVE   & FPS  \\
    \midrule
     OA-SparseBEV & 43.1  &   53.4   &    0.638   & 0.463 & 0.257 &  12.6   \\ 
     +LightW. Opti. & 43.6  & 53.5 & 0.624   & 0.481 & 0.263 & 13.5 \\
     +IPTE   & 43.9 & 53.9 & 0.638  & 0.452 & 0.255 & 13.5 \\

    \bottomrule
    \end{tabular}
  \label{tab:temp2d}
\end{table}

\subsection{Temporal Optimization for Object-Aware Branch}

To enhance temporal information processing within our plug-and-play framework while maintaining deployment efficiency, we implement two synergistic optimizations. First, we introduce lightweight modifications to the 2D detection heads, \emph{e.g.}, FCOS or YoloX head, to address computational redundancy in multi-frame inputs. While retaining multi-scale pyramid features $\mathcal{M}_t = \{F_t^8, F_t^{16}, F_t^{32}, F_t^{64}\}$ for key frame ($\tau_t$) processing during training to enable scale-specific foreground selection, we unify the feature resolution to 16$\times$ down-sampling for full-scale detection in both training and inference phases. For historical frames ($\hat{\tau}$), we exclusively adopt single-layer 16$\times$ down-sampled features through a simplified FPN, achieving computational efficiency without compromising detection capability. This design leverages the observation that temporal influence diminishes with frame distance, while maintaining critical spatial details in the current frame.

Second, we develop an image-view predictive temporal enhancement (IPTE) module to strengthen temporal consistency during training. 
Previous studies~\cite{hop,pred2det} have demonstrated that multi-frame fusion and the prediction of intermediate or future frames on BEV features can enhance the temporal modeling capability of models, thereby improving detection accuracy.
Unlike temporal enhancement on BEV, which explicitly utilizes camera intrinsic and extrinsic parameters, the image-view enhancement inherently encodes camera intrinsics and ego-motion.
The IPTE processes historical frames $\{\hat{\tau}_{t-k},...,\hat{\tau}_{t-1}\}$ through deformable attention-based feature fusion:
\begin{flalign}
    &&
    F^p_{t} = \text{DeformAttn}(\{F_{t-i}^{16}\}_{i=1}^k),
    &&
\end{flalign}
where fused features $F^p_t$ drive parallel predictions of 2D bounding boxes and depths for the key frame. Crucially, this auxiliary module is deactivated during inference, ensuring zero additional computational overhead.

\parsection{Experimental Validation}
We evaluate the temporal optimizations on the nuScenes validation set using four consecutive frames as input, as shown in Table~\ref{tab:temp2d}. The baseline OA-SparseBEV achieves a base performance of 43.1\% mAP and 53.4\% NDS with an inference speed of 12.6 FPS. After implementing the lightweight optimization strategy for 2D detection, the model exhibits a 7.1\% improvement in frame rate, reaching 13.5 FPS, while simultaneously increasing mAP by 0.5 percentage points and reducing mean absolute translation error (mATE) by 0.014. This improvement is attributed to the enhanced global perception capability enabled by constraining the 2D detection head to unified 16$\times$ down-sampled features.

Subsequently, integrating the IPTE module with two historical frames further improves performance, achieving a final mAP of 43.9\% and NDS of 53.9\%. This represents an overall improvement of 0.8\% in mAP and 0.5\% in NDS compared to the baseline. 
Crucially, these enhancements are accomplished without introducing additional latency during inference, thereby preserving the framework’s deployment efficiency.

\subsection{Analysis of Effect on Different Metrics}
\systemname outperforms most baseline models in terms of three static attributes of individual objects: mean Absolute Translation Error (mATE), mean Absolute Scale Error (mASE), and mean Average Precision (mAP). These metrics measure the accuracy of object localization, scale estimation, and overall detection performance, respectively. \systemname demonstrates superior performance compared to the baselines in these aspects.
However, when it comes to metrics that specifically capture the motion-related characteristics of objects, such as mean Absolute Velocity Error (mAVE) and mean Absolute Angular Error (mAAE), the improvements achieved by \systemname are not as pronounced. The reason behind this phenomenon lies in the emphasis of our method on foreground object awareness, which primarily enhances the classification of objects rather than improving their motion prediction capabilities. As a result, the advancements made by our method in metrics related to object motion are relatively modest.
In future work, the focus will shift towards exploring object awareness across consecutive frames. By considering the information from multiple frames and incorporating temporal context, it is expected that the motion prediction performance of \systemname can be further enhanced.

\subsection{Comprehensive Analysis}

\systemname systematically integrates object-aware depth supervision and foreground pseudo-3D representation as a general plug-in module, enabling robust and adaptable performance across diverse 3D detection frameworks. By leveraging 3D bounding box properties, \emph{e.g.}, center points or diagonal slices, to generate pseudo-depth labels, the method effectively mitigates depth ambiguity inherent in view transformation while avoiding reliance on additional annotations. This object-centric depth estimation is further enhanced by projecting 2D foreground pixels into sparse pseudo-LiDAR point clouds, which are encoded via 3D sparse convolutions to capture precise spatial structures.

To rigorously validate the robustness of our results, extensive validation across nuScenes and Argoverse 2 datasets demonstrates consistent improvements in both BEV-based and query-based models, particularly under challenging conditions such as nighttime, rainy weather, and long-range detection, such as a 9.6\% mAP gain in rainy conditions and a 16.4\% mAP increase for distant objects. 
While \systemname introduces a moderate computational overhead due to the integration of object-aware modules, our design strategically balances performance gains with practical deployability. 
By adopting pillar-based encoding and lightweight 2D detection heads, we constrain parameter increases while achieving notable performance gain. 
For example, with our lightweight optimization and temporal enhancement for the object-aware plugin, OA-SparseBEV using 4 frames achieves a 1.9\% relative improvement in mAP while also increasing inference speed by 7.1\% in FPS. 
Future work will further optimize runtime through GPU-accelerated pseudo-point generation and quantization.

\section{Conclusion and Discussion}
\label{sec:conclu}

In this paper, we propose \systemname, a method that can be plugged into multi-camera 3D object detection frameworks. The key innovation of \systemname is utilizing object-aware features to distinguish objects from the background, thereby mitigating deformation effects resulting from view transformation.
In other words, \systemname utilizes object-level supervision to estimate the object depth map and extract foreground pseudo-3D features that indicate an object's position and 3D structure. Integrating these features into the BEV representation enhances their descriptive capabilities for objects of interest and effectively improves detection performance. We implement our method in multiple representative baseline networks and demonstrate its effectiveness through consistent improvements on both the nuScenes dataset and the Argoverse 2 dataset.

The core strength of \systemname lies in its generalizability as a plug-in module, enabling seamless integration with BEV-based and query-based pipelines while avoiding reliance on additional annotations. By explicitly incorporating object-centric depth and pseudo-3D features, it addresses inherent ambiguities in view transformation, particularly for long-range detection and cluttered scenes. However, two limitations warrant discussion: (i) The current implementation introduces a computational overhead ($\sim$12–15\% latency increase) due to pseudo-voxel encoding and 2D object detection. The ongoing optimizations, \emph{e.g.}, the pillar-based 3D feature encoding and lightweight 2D detectors, aim to mitigate this. (ii) Performance partly depends on the accuracy of the 2D object detector, which may struggle in low-visibility scenarios. 
Future work will explore the fusion method with radars or infrared cameras to reduce this dependency. These limitations do not diminish the core contribution of our method, but highlight opportunities for refinement in real-time and robust applications.

\parsection{Data availability} 
The nuScenes~\citep{nuScenes} dataset and the Argoverse 2~\citep{av2} dataset can be obtained from \url{https://www.nuscenes.org/} and \url{https://www.argoverse.org/av2.html}, respectively. The codes that support the findings of this study are available from the corresponding author, \emph{i.e.}, Yanyong Zhang and Jiajun Deng, upon reasonable request.

\section*{Declarations}

\textbf{Conflict of interest} There are no conflicts to declare.


\bibliographystyle{spbasic}

\bibliography{sn-bibliography}

\begin{thebibliography}{66}
\providecommand{\natexlab}[1]{#1}
\providecommand{\url}[1]{{#1}}
\providecommand{\urlprefix}{URL }
\expandafter\ifx\csname urlstyle\endcsname\relax
  \providecommand{\doi}[1]{DOI~\discretionary{}{}{}#1}\else
  \providecommand{\doi}{DOI~\discretionary{}{}{}\begingroup \urlstyle{rm}\Url}\fi
\providecommand{\eprint}[2][]{\url{#2}}

\bibitem[{Brazil and Liu(2019)}]{M3D-RPN}
Brazil G, Liu X (2019) M3d-rpn: Monocular 3d region proposal network for object detection. In: Proceedings of the IEEE/CVF international conference on computer vision, pp 9287--9296

\bibitem[{Caesar et~al.(2020)Caesar, Bankiti, Lang, Vora, Liong, Xu, Krishnan, Pan, Baldan, and Beijbom}]{nuScenes}
Caesar H, Bankiti V, Lang AH, Vora S, Liong VE, Xu Q, Krishnan A, Pan Y, Baldan G, Beijbom O (2020) nuscenes: A multimodal dataset for autonomous driving. In: Proceedings of the IEEE/CVF conference on computer vision and pattern recognition, pp 11621--11631

\bibitem[{Carion et~al.(2020)Carion, Massa, Synnaeve, Usunier, Kirillov, and Zagoruyko}]{DETR}
Carion N, Massa F, Synnaeve G, Usunier N, Kirillov A, Zagoruyko S (2020) End-to-end object detection with transformers. In: European conference on computer vision, Springer, pp 213--229

\bibitem[{Chu et~al.(2021)Chu, Deng, Li, Yuan, Zhang, Ji, and Zhang}]{Neighbor-Vote}
Chu X, Deng J, Li Y, Yuan Z, Zhang Y, Ji J, Zhang Y (2021) Neighbor-vote: Improving monocular 3d object detection through neighbor distance voting. In: Proceedings of the 29th ACM International Conference on Multimedia, pp 5239--5247

\bibitem[{Dai et~al.(2017)Dai, Qi, Xiong, Li, Zhang, Hu, and Wei}]{DCN}
Dai J, Qi H, Xiong Y, Li Y, Zhang G, Hu H, Wei Y (2017) Deformable convolutional networks. In: Proceedings of the IEEE international conference on computer vision, pp 764--773

\bibitem[{Deng et~al.(2021)Deng, Shi, Li, Zhou, Zhang, and Li}]{Voxel-R-CNN}
Deng J, Shi S, Li P, Zhou W, Zhang Y, Li H (2021) Voxel r-cnn: Towards high performance voxel-based 3d object detection. In: Proceedings of the AAAI conference on artificial intelligence, vol~35, pp 1201--1209

\bibitem[{Ding et~al.(2020)Ding, Huo, Yi, Wang, Shi, Lu, and Luo}]{D4LCN}
Ding M, Huo Y, Yi H, Wang Z, Shi J, Lu Z, Luo P (2020) Learning depth-guided convolutions for monocular 3d object detection. In: Proceedings of the IEEE/CVF Conference on computer vision and pattern recognition workshops, pp 1000--1001

\bibitem[{Fu et~al.(2018)Fu, Gong, Wang, Batmanghelich, and Tao}]{DORN}
Fu H, Gong M, Wang C, Batmanghelich K, Tao D (2018) Deep ordinal regression network for monocular depth estimation. In: Proceedings of the IEEE conference on computer vision and pattern recognition, pp 2002--2011

\bibitem[{Graham(2015)}]{sparse-3D-conv}
Graham B (2015) Sparse 3d convolutional neural networks. In: Xie X, Jones MW, Tam GKL (eds) Proceedings of the British Machine Vision Conference 2015, pp 150.1--150.9

\bibitem[{Guizilini et~al.(2020)Guizilini, Ambrus, Pillai, Raventos, and Gaidon}]{DDAD}
Guizilini V, Ambrus R, Pillai S, Raventos A, Gaidon A (2020) 3d packing for self-supervised monocular depth estimation. In: Proceedings of the IEEE/CVF conference on computer vision and pattern recognition, pp 2485--2494

\bibitem[{Han et~al.(2024)Han, Yang, Sun, Ge, Dong, Zhou, Mao, Peng, and Zhang}]{videobev}
Han C, Yang J, Sun J, Ge Z, Dong R, Zhou H, Mao W, Peng Y, Zhang X (2024) Exploring recurrent long-term temporal fusion for multi-view 3d perception. {IEEE} Robotics Autom Lett 9(7):6544--6551

\bibitem[{He et~al.(2016)He, Zhang, Ren, and Sun}]{ResNet}
He K, Zhang X, Ren S, Sun J (2016) Deep residual learning for image recognition. In: Proceedings of the IEEE conference on computer vision and pattern recognition, pp 770--778

\bibitem[{Huang and Huang(2022)}]{BEVDet4D}
Huang J, Huang G (2022) Bevdet4d: Exploit temporal cues in multi-camera 3d object detection. arXiv preprint arXiv:220317054

\bibitem[{Huang et~al.(2021)Huang, Huang, Zhu, Ye, and Du}]{BEVDet}
Huang J, Huang G, Zhu Z, Ye Y, Du D (2021) Bevdet: High-performance multi-camera 3d object detection in bird-eye-view. arXiv preprint arXiv:211211790

\bibitem[{Jiang et~al.(2024)Jiang, Li, Liu, Wang, Jia, Wang, Han, and Zhang}]{far3d}
Jiang X, Li S, Liu Y, Wang S, Jia F, Wang T, Han L, Zhang X (2024) Far3d: Expanding the horizon for surround-view 3d object detection. In: Proceedings of the AAAI Conference on Artificial Intelligence, vol~38, pp 2561--2569

\bibitem[{Jiang et~al.(2023)Jiang, Zhang, Miao, Zhu, Gao, Hu, and Jiang}]{PolarFormer}
Jiang Y, Zhang L, Miao Z, Zhu X, Gao J, Hu W, Jiang YG (2023) Polarformer: Multi-camera 3d object detection with polar transformer. In: Proceedings of the AAAI conference on Artificial Intelligence, vol~37, pp 1042--1050

\bibitem[{Kim et~al.(2023)Kim, Kim, Lee, and Kum}]{pred2det}
Kim S, Kim Y, Lee IJ, Kum D (2023) Predict to detect: Prediction-guided 3d object detection using sequential images. In: Proceedings of the IEEE/CVF International Conference on Computer Vision, pp 18057--18066

\bibitem[{Lee et~al.(2019{\natexlab{a}})Lee, Hwang, Lee, Bae, and Park}]{V2-99}
Lee Y, Hwang Jw, Lee S, Bae Y, Park J (2019{\natexlab{a}}) An energy and gpu-computation efficient backbone network for real-time object detection. In: Proceedings of the IEEE/CVF conference on computer vision and pattern recognition workshops, pp 0--0

\bibitem[{Lee et~al.(2019{\natexlab{b}})Lee, Hwang, Lee, Bae, and Park}]{vovnet}
Lee Y, Hwang Jw, Lee S, Bae Y, Park J (2019{\natexlab{b}}) An energy and gpu-computation efficient backbone network for real-time object detection. In: Proceedings of the IEEE/CVF conference on computer vision and pattern recognition workshops, pp 0--0

\bibitem[{Li et~al.(2020)Li, Zhao, Liu, and Cao}]{RTM3D}
Li P, Zhao H, Liu P, Cao F (2020) Rtm3d: Real-time monocular 3d detection from object keypoints for autonomous driving. In: European Conference on Computer Vision, Springer, pp 644--660

\bibitem[{Li et~al.(2022)Li, Chen, Qi, Li, Sun, and Jia}]{UVTR}
Li Y, Chen Y, Qi X, Li Z, Sun J, Jia J (2022) Unifying voxel-based representation with transformer for 3d object detection. Advances in Neural Information Processing Systems 35:18442--18455

\bibitem[{Li et~al.(2023{\natexlab{a}})Li, Bao, Ge, Yang, Sun, and Li}]{BEVStereo}
Li Y, Bao H, Ge Z, Yang J, Sun J, Li Z (2023{\natexlab{a}}) Bevstereo: Enhancing depth estimation in multi-view 3d object detection with temporal stereo. In: Proceedings of the AAAI Conference on Artificial Intelligence, vol~37, pp 1486--1494

\bibitem[{Li et~al.(2023{\natexlab{b}})Li, Ge, Yu, Yang, Wang, Shi, Sun, and Li}]{BEVDepth}
Li Y, Ge Z, Yu G, Yang J, Wang Z, Shi Y, Sun J, Li Z (2023{\natexlab{b}}) Bevdepth: Acquisition of reliable depth for multi-view 3d object detection. In: Proceedings of the AAAI conference on artificial intelligence, vol~37, pp 1477--1485

\bibitem[{Li et~al.(2023{\natexlab{c}})Li, Yu, Wang, Anandkumar, Lu, and Alvarez}]{FB-BEV}
Li Z, Yu Z, Wang W, Anandkumar A, Lu T, Alvarez JM (2023{\natexlab{c}}) Fb-bev: Bev representation from forward-backward view transformations. In: Proceedings of the IEEE/CVF International Conference on Computer Vision, pp 6919--6928

\bibitem[{Li et~al.(2024)Li, Wang, Li, Xie, Sima, Lu, Yu, and Dai}]{BEVFormer}
Li Z, Wang W, Li H, Xie E, Sima C, Lu T, Yu Q, Dai J (2024) Bevformer: learning bird's-eye-view representation from lidar-camera via spatiotemporal transformers. IEEE Transactions on Pattern Analysis and Machine Intelligence

\bibitem[{Lin et~al.(2017)Lin, Doll{\'a}r, Girshick, He, Hariharan, and Belongie}]{FPN}
Lin TY, Doll{\'a}r P, Girshick R, He K, Hariharan B, Belongie S (2017) Feature pyramid networks for object detection. In: Proceedings of the IEEE conference on computer vision and pattern recognition, pp 2117--2125

\bibitem[{Lin et~al.(2022)Lin, Lin, Pei, Huang, and Su}]{sparse4d}
Lin X, Lin T, Pei Z, Huang L, Su Z (2022) Sparse4d: Multi-view 3d object detection with sparse spatial-temporal fusion. arXiv preprint arXiv:221110581

\bibitem[{Lin et~al.(2023{\natexlab{a}})Lin, Lin, Pei, Huang, and Su}]{Sparse4Dv2}
Lin X, Lin T, Pei Z, Huang L, Su Z (2023{\natexlab{a}}) Sparse4d v2: Recurrent temporal fusion with sparse model. arXiv preprint arXiv:230514018

\bibitem[{Lin et~al.(2023{\natexlab{b}})Lin, Pei, Lin, Huang, and Su}]{sparse4dv3}
Lin X, Pei Z, Lin T, Huang L, Su Z (2023{\natexlab{b}}) Sparse4d v3: Advancing end-to-end 3d detection and tracking. CoRR abs/2311.11722

\bibitem[{Liu et~al.(2024)Liu, Huang, Zhang, Yao, Zhang, Wan, Ye, and Zhou}]{RayDN}
Liu F, Huang T, Zhang Q, Yao H, Zhang C, Wan F, Ye Q, Zhou Y (2024) Ray denoising: Depth-aware hard negative sampling for multi-view 3d object detection. In: European Conference on Computer Vision, Springer, pp 200--217

\bibitem[{Liu et~al.(2023{\natexlab{a}})Liu, Teng, Lu, Wang, and Wang}]{sparsebev}
Liu H, Teng Y, Lu T, Wang H, Wang L (2023{\natexlab{a}}) Sparsebev: High-performance sparse 3d object detection from multi-camera videos. In: Proceedings of the IEEE/CVF International Conference on Computer Vision, pp 18580--18590

\bibitem[{Liu et~al.(2022)Liu, Wang, Zhang, and Sun}]{PETR}
Liu Y, Wang T, Zhang X, Sun J (2022) Petr: Position embedding transformation for multi-view 3d object detection. In: European conference on computer vision, Springer, pp 531--548

\bibitem[{Liu et~al.(2023{\natexlab{b}})Liu, Yan, Jia, Li, Gao, Wang, and Zhang}]{petrv2}
Liu Y, Yan J, Jia F, Li S, Gao A, Wang T, Zhang X (2023{\natexlab{b}}) Petrv2: A unified framework for 3d perception from multi-camera images. In: Proceedings of the IEEE/CVF International Conference on Computer Vision, pp 3262--3272

\bibitem[{Liu et~al.(2020)Liu, Wu, and T{\'o}th}]{SMOKE}
Liu Z, Wu Z, T{\'o}th R (2020) Smoke: Single-stage monocular 3d object detection via keypoint estimation. In: Proceedings of the IEEE/CVF conference on computer vision and pattern recognition workshops, pp 996--997

\bibitem[{Liu et~al.(2021)Liu, Lin, Cao, Hu, Wei, Zhang, Lin, and Guo}]{ST}
Liu Z, Lin Y, Cao Y, Hu H, Wei Y, Zhang Z, Lin S, Guo B (2021) Swin transformer: Hierarchical vision transformer using shifted windows. In: Proceedings of the IEEE/CVF international conference on computer vision, pp 10012--10022

\bibitem[{Loshchilov and Hutter(2017)}]{cosinedecay}
Loshchilov I, Hutter F (2017) {SGDR:} stochastic gradient descent with warm restarts. In: International Conference on Learning Representations ({ICLR})

\bibitem[{Loshchilov and Hutter(2019)}]{adamw}
Loshchilov I, Hutter F (2019) Decoupled weight decay regularization. In: International Conference on Learning Representations ({ICLR})

\bibitem[{Ma et~al.(2019)Ma, Wang, Li, Zhang, Ouyang, and Fan}]{AM3D}
Ma X, Wang Z, Li H, Zhang P, Ouyang W, Fan X (2019) Accurate monocular 3d object detection via color-embedded 3d reconstruction for autonomous driving. In: Proceedings of the IEEE/CVF international conference on computer vision, pp 6851--6860

\bibitem[{Ma et~al.(2020)Ma, Liu, Xia, Zhang, Zeng, and Ouyang}]{patchnet}
Ma X, Liu S, Xia Z, Zhang H, Zeng X, Ouyang W (2020) Rethinking pseudo-lidar representation. In: Computer Vision--ECCV 2020: 16th European Conference, Glasgow, UK, August 23--28, 2020, Proceedings, Part XIII 16, Springer, pp 311--327

\bibitem[{Ma et~al.(2024)Ma, Ouyang, Simonelli, and Ricci}]{3dod_tpami_survey}
Ma X, Ouyang W, Simonelli A, Ricci E (2024) 3d object detection from images for autonomous driving: {A} survey. {IEEE} Trans Pattern Anal Mach Intell 46(5):3537--3556

\bibitem[{Mao et~al.(2023)Mao, Shi, Wang, and Li}]{3dod_ijcv_survey1}
Mao J, Shi S, Wang X, Li H (2023) 3d object detection for autonomous driving: {A} comprehensive survey. Int J Comput Vis 131(8):1909--1963

\bibitem[{Mousavian et~al.(2017)Mousavian, Anguelov, Flynn, and Kosecka}]{Deep3DBox}
Mousavian A, Anguelov D, Flynn J, Kosecka J (2017) 3d bounding box estimation using deep learning and geometry. In: Proceedings of the IEEE conference on Computer Vision and Pattern Recognition, pp 7074--7082

\bibitem[{Park et~al.(2021)Park, Ambrus, Guizilini, Li, and Gaidon}]{DD3D}
Park D, Ambrus R, Guizilini V, Li J, Gaidon A (2021) Is pseudo-lidar needed for monocular 3d object detection? In: Proceedings of the IEEE/CVF International Conference on Computer Vision, pp 3142--3152

\bibitem[{Park et~al.(2023)Park, Xu, Yang, Keutzer, Kitani, Tomizuka, and Zhan}]{SOLOFusion}
Park J, Xu C, Yang S, Keutzer K, Kitani KM, Tomizuka M, Zhan W (2023) Time will tell: New outlooks and {A} baseline for temporal multi-view 3d object detection. In: International Conference on Learning Representations, {ICLR}

\bibitem[{Philion and Fidler(2020)}]{LSS}
Philion J, Fidler S (2020) Lift, splat, shoot: Encoding images from arbitrary camera rigs by implicitly unprojecting to 3d. In: Computer Vision--ECCV 2020: 16th European Conference, Glasgow, UK, August 23--28, 2020, Proceedings, Part XIV 16, Springer, pp 194--210

\bibitem[{Qi et~al.(2017{\natexlab{a}})Qi, Su, Mo, and Guibas}]{pointnet}
Qi CR, Su H, Mo K, Guibas LJ (2017{\natexlab{a}}) Pointnet: Deep learning on point sets for 3d classification and segmentation. In: Proceedings of the IEEE conference on computer vision and pattern recognition, pp 652--660

\bibitem[{Qi et~al.(2017{\natexlab{b}})Qi, Yi, Su, and Guibas}]{pointnet++}
Qi CR, Yi L, Su H, Guibas LJ (2017{\natexlab{b}}) Pointnet++: Deep hierarchical feature learning on point sets in a metric space. Advances in neural information processing systems 30

\bibitem[{Reading et~al.(2021)Reading, Harakeh, Chae, and Waslander}]{CaDDN}
Reading C, Harakeh A, Chae J, Waslander SL (2021) Categorical depth distribution network for monocular 3d object detection. In: Proceedings of the IEEE/CVF conference on computer vision and pattern recognition, pp 8555--8564

\bibitem[{Shi et~al.(2023)Shi, Jiang, Deng, Wang, Guo, Shi, Wang, and Li}]{PV-RCNN++}
Shi S, Jiang L, Deng J, Wang Z, Guo C, Shi J, Wang X, Li H (2023) Pv-rcnn++: Point-voxel feature set abstraction with local vector representation for 3d object detection. International Journal of Computer Vision 131(2):531--551

\bibitem[{Tian et~al.(2019)Tian, Shen, Chen, and He}]{FCOS}
Tian Z, Shen C, Chen H, He T (2019) Fcos: Fully convolutional one-stage object detection. In: Proceedings of the IEEE/CVF international conference on computer vision, pp 9627--9636

\bibitem[{Vaswani et~al.(2017)Vaswani, Shazeer, Parmar, Uszkoreit, Jones, Gomez, Kaiser, and Polosukhin}]{Attention}
Vaswani A, Shazeer N, Parmar N, Uszkoreit J, Jones L, Gomez AN, Kaiser {\L}, Polosukhin I (2017) Attention is all you need. Advances in neural information processing systems 30

\bibitem[{Wang et~al.(2023{\natexlab{a}})Wang, Liu, Wang, Li, and Zhang}]{streampetr}
Wang S, Liu Y, Wang T, Li Y, Zhang X (2023{\natexlab{a}}) Exploring object-centric temporal modeling for efficient multi-view 3d object detection. In: Proceedings of the IEEE/CVF international conference on computer vision, pp 3621--3631

\bibitem[{Wang et~al.(2021)Wang, Zhu, Pang, and Lin}]{FCOS3D}
Wang T, Zhu X, Pang J, Lin D (2021) Fcos3d: Fully convolutional one-stage monocular 3d object detection. In: Proceedings of the IEEE/CVF international conference on computer vision, pp 913--922

\bibitem[{Wang et~al.(2019)Wang, Chao, Garg, Hariharan, Campbell, and Weinberger}]{Pseudo-Lidar}
Wang Y, Chao WL, Garg D, Hariharan B, Campbell M, Weinberger KQ (2019) Pseudo-lidar from visual depth estimation: Bridging the gap in 3d object detection for autonomous driving. In: Proceedings of the IEEE/CVF conference on computer vision and pattern recognition, pp 8445--8453

\bibitem[{Wang et~al.(2022)Wang, Guizilini, Zhang, Wang, Zhao, and Solomon}]{DETR3D}
Wang Y, Guizilini VC, Zhang T, Wang Y, Zhao H, Solomon J (2022) Detr3d: 3d object detection from multi-view images via 3d-to-2d queries. In: Conference on Robot Learning, PMLR, pp 180--191

\bibitem[{Wang et~al.(2023{\natexlab{b}})Wang, Mao, Zhu, Deng, Zhang, Ji, Li, and Zhang}]{3dod_ijcv_survey2}
Wang Y, Mao Q, Zhu H, Deng J, Zhang Y, Ji J, Li H, Zhang Y (2023{\natexlab{b}}) Multi-modal 3d object detection in autonomous driving: {A} survey. Int J Comput Vis 131(8):2122--2152

\bibitem[{Wang et~al.(2023{\natexlab{c}})Wang, Huang, Fu, Wang, and Liu}]{mv2d}
Wang Z, Huang Z, Fu J, Wang N, Liu S (2023{\natexlab{c}}) Object as query: Lifting any 2d object detector to 3d detection. In: Proceedings of the IEEE/CVF International Conference on Computer Vision, pp 3791--3800

\bibitem[{Wilson et~al.(2021)Wilson, Qi, Agarwal, Lambert, Singh, Khandelwal, Pan, Kumar, Hartnett, Pontes, Ramanan, Carr, and Hays}]{av2}
Wilson B, Qi W, Agarwal T, Lambert J, Singh J, Khandelwal S, Pan B, Kumar R, Hartnett A, Pontes JK, Ramanan D, Carr P, Hays J (2021) Argoverse 2: Next generation datasets for self-driving perception and forecasting. In: Proceedings of the Neural Information Processing Systems Track on Datasets and Benchmarks 1, NeurIPS Datasets and Benchmarks 2021, December 2021, virtual

\bibitem[{Yin et~al.(2021)Yin, Zhou, and Krahenbuhl}]{CenterPoint3D}
Yin T, Zhou X, Krahenbuhl P (2021) Center-based 3d object detection and tracking. In: Proceedings of the IEEE/CVF conference on computer vision and pattern recognition, pp 11784--11793

\bibitem[{Zablocki et~al.(2022)Zablocki, Ben{-}Younes, P{\'{e}}rez, and Cord}]{auto_driving_survey1}
Zablocki {\'{E}}, Ben{-}Younes H, P{\'{e}}rez P, Cord M (2022) Explainability of deep vision-based autonomous driving systems: Review and challenges. Int J Comput Vis 130(10):2425--2452

\bibitem[{Zhang et~al.(2023)Zhang, Qiu, Wang, Guo, Cui, Qiao, Li, and Gao}]{MonoDETR}
Zhang R, Qiu H, Wang T, Guo Z, Cui Z, Qiao Y, Li H, Gao P (2023) Monodetr: Depth-guided transformer for monocular 3d object detection. In: Proceedings of the IEEE/CVF International Conference on Computer Vision, pp 9155--9166

\bibitem[{Zhang et~al.(2021)Zhang, Lu, and Zhou}]{MonoFlex}
Zhang Y, Lu J, Zhou J (2021) Objects are different: Flexible monocular 3d object detection. In: Proceedings of the IEEE/CVF Conference on Computer Vision and Pattern Recognition, pp 3289--3298

\bibitem[{Zhang et~al.(2022)Zhang, Zhu, Zheng, Huang, Huang, Zhou, and Lu}]{BEVerse}
Zhang Y, Zhu Z, Zheng W, Huang J, Huang G, Zhou J, Lu J (2022) Beverse: Unified perception and prediction in birds-eye-view for vision-centric autonomous driving. arXiv preprint arXiv:220509743

\bibitem[{Zhu et~al.(2019)Zhu, Jiang, Zhou, Li, and Yu}]{CBGS}
Zhu B, Jiang Z, Zhou X, Li Z, Yu G (2019) Class-balanced grouping and sampling for point cloud 3d object detection. arXiv preprint arXiv:190809492

\bibitem[{Zhu et~al.(2021)Zhu, Su, Lu, Li, Wang, and Dai}]{Deformable-DETR}
Zhu X, Su W, Lu L, Li B, Wang X, Dai J (2021) Deformable {DETR:} deformable transformers for end-to-end object detection. In: International Conference on Learning Representations ({ICLR})

\bibitem[{Zong et~al.(2023)Zong, Jiang, Song, Xue, Su, Li, and Liu}]{hop}
Zong Z, Jiang D, Song G, Xue Z, Su J, Li H, Liu Y (2023) Temporal enhanced training of multi-view 3d object detector via historical object prediction. In: Proceedings of the IEEE/CVF International Conference on Computer Vision, pp 3781--3790

\end{thebibliography}

\end{document}